\documentclass{article}

\usepackage[preprint]{neurips_2023}

\usepackage{amsmath,amsfonts,bm}

\def\eqref#1{equation~\ref{#1}}
\def\1{\bm{1}}

\DeclareMathAlphabet{\mathsfit}{\encodingdefault}{\sfdefault}{m}{sl}
\SetMathAlphabet{\mathsfit}{bold}{\encodingdefault}{\sfdefault}{bx}{n}

\usepackage[utf8]{inputenc} % allow utf-8 input
\usepackage[T1]{fontenc}    % use 8-bit T1 fonts
\usepackage{hyperref}       % hyperlinks
\usepackage{url}            % simple URL typesetting
\usepackage{booktabs}       % professional-quality tables
\usepackage{amsfonts}       % blackboard math symbols
\usepackage{nicefrac}       % compact symbols for 1/2, etc.
\usepackage{microtype}      % microtypography
\usepackage{xcolor}         % colors

\hypersetup{
    colorlinks,
    linkcolor={red!50!black},
    citecolor={blue!50!black},
    urlcolor={blue!80!black}
}
\usepackage{amsmath}
\usepackage{pgfplots}
\usepackage{pgfplotstable}
\usepackage{graphicx}
\usepackage{subcaption}
\usepackage{tikz}
\usetikzlibrary{pgfplots.groupplots}
\usetikzlibrary{decorations}
\usetikzlibrary{decorations.markings}
\usetikzlibrary{arrows.meta}
\usepackage{ragged2e}
\usepackage{booktabs}
\usepackage[capitalise]{cleveref}
\usepackage[normalem]{ulem}
\usepackage{algorithm}
\usepackage{algpseudocode}
\usepackage{multirow}

\newcommand{\ie}{\textit{i.e.} }

\title{A Simple and Efficient Baseline for \\Data Attribution on Images}

\author{Vasu~Singla$^{1}$ \quad Pedro~Sandoval-Segura$^{1}$ \quad Micah~Goldblum$^{2}$ \quad \\
\textbf{Jonas~Geiping$^{3}$ \quad
Tom~Goldstein}$^{1}$ \\
$^{1}$University of Maryland \quad $^{2}$New York University \\
$^{3}$ ELLIS Institute Tübingen, MPI for Intelligent Systems, Tübingen AI Center \\
{\tt\small \{vsingla, psando, tomg\}@cs.umd.edu \quad \tt\small jonas@tue.ellis.edu \quad \tt\small goldblum@nyu.edu }
}

\begin{document}

\maketitle

\begin{abstract}
Data attribution methods play a crucial role in understanding machine learning models, providing insight into which training data points are most responsible for model outputs during deployment. However, current state-of-the-art approaches require a large ensemble of as many as 300,000 models to accurately attribute model predictions. These approaches therefore come at a high computational cost, are memory intensive, and are hard to scale to large models or datasets. 
In this work, we focus on a minimalist baseline that relies on the image features from a pretrained self-supervised backbone to retrieve images from the dataset. Our method is model-agnostic and scales easily to large datasets. We show results on CIFAR-10 and ImageNet, achieving strong performance that rivals or outperforms state-of-the-art approaches at a fraction of the compute or memory cost. Contrary to prior work, our results reinforce the intuition that a model's prediction on one image is most impacted by visually similar training samples. Our approach serves as a simple and efficient baseline for data attribution on images.
% Thoughts about ending like this?
% Other endings @Jonas - This provides critical insight into the limits of progress in the field of data attribution. Consider for arxiv

% Our method is model agnostic and focuses on simplicity, and efficiency while scaling easily to large datasets. 

% and demonstrates that strong attributions can be obtained without large model ensembles and at low computational cost.  

% One option:
% The ability to trace a neural network's predictions back to its training data is an important step in understanding a network's errors, biases, and capabilities. But approximating the end-to-end training and evaluation of deep neural networks is extremely complex, and the most effective methods have involved training between hundreds to millions of models. In this work, we propose a method that only involves training a single model. Our work reinforces the intuition that visual similarity plays an important role in data attribution
\end{abstract}

\section{Introduction}
\label{sec:intro}

% Motivate by problem first, then describe data attribution approach as the solution. Issues of data attribution
% Influence function Language Model Call Out, Anthropic 
% Safety critical for Industry
% 
The effectiveness of a machine learning system's performance hinges on the quality, diversity, and relevance of the data it is trained on \citep{halevy2009unreasonable, sun2017revisiting}.  In various real-world machine learning systems, for example in healthcare or finance, we often ask questions like, ``Which training samples influenced this prediction?" or ``How sensitive is this model's prediction to changes in the training data?" Counterfactual insights enable us to assess the impact of hypothetical changes in the data distribution, which in turn helps us understand the basis of the model's decisions and how to change the decision in the event of an error.

These questions motivate research on \emph{data attribution} methods, which focus on understanding which data points most strongly influence a model's outputs. Data attribution methods have been applied to applications such as debugging model biases \citep{ilyas22datamodels, park2023trak, shah2023modeldiff},  fairness assessment \citep{black2021leave}, and active learning~\citep{liu2021active}. 

\begin{figure}
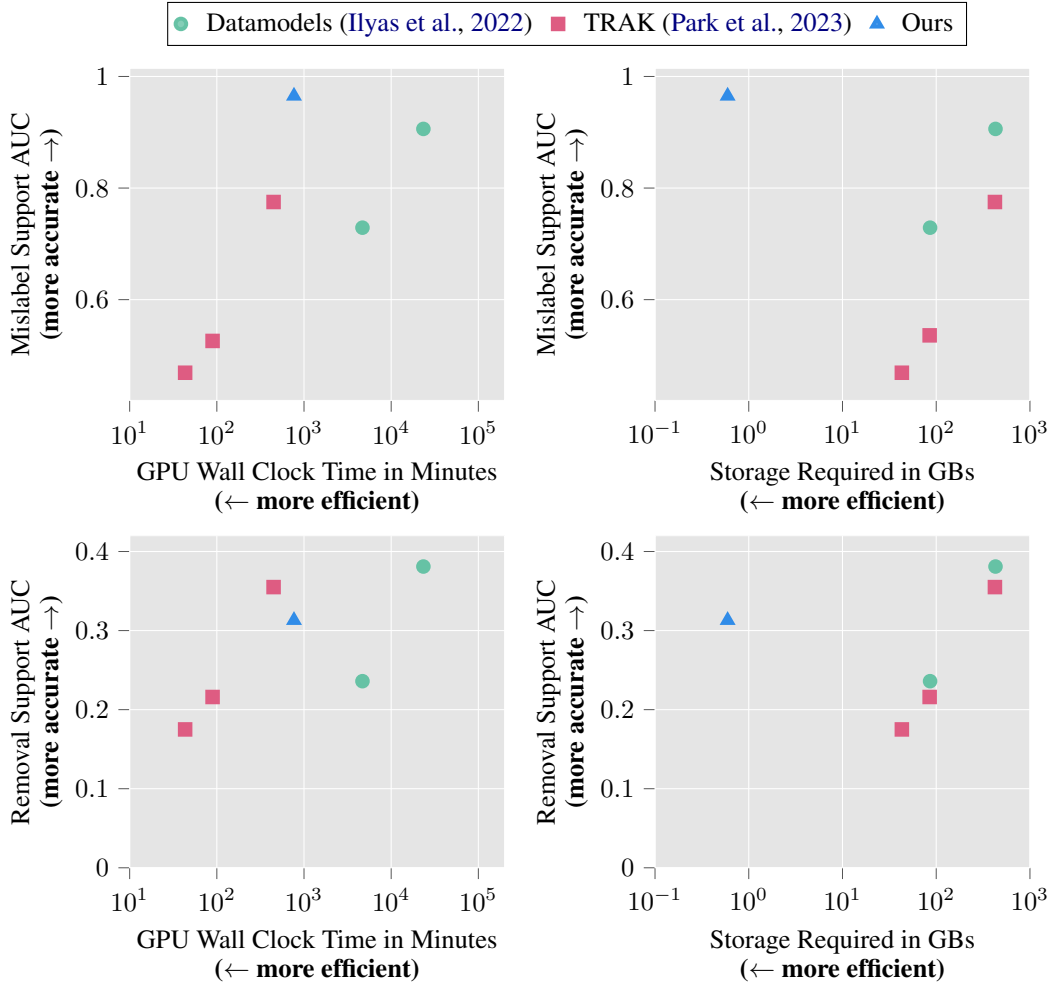

    \centering
\begin{tikzpicture}
  \centering
    \definecolor{darkgray141160203}{RGB}{141,160,203}
    \definecolor{dimgray85}{RGB}{85,85,85}
    \definecolor{gainsboro229}{RGB}{229,229,229}
    \definecolor{lightgray204}{RGB}{204,204,204}
    \definecolor{mediumaquamarine102194165}{RGB}{102,194,165}
    \definecolor{orchid231138195}{RGB}{231,138,195}
    \definecolor{salmon25214198}{RGB}{252,141,98}
    \definecolor{amethyst}{rgb}{0.6, 0.4, 0.8}
    \definecolor{bleudefrance}{rgb}{0.19, 0.55, 0.91}
    \definecolor{blush}{rgb}{0.87, 0.36, 0.51}
    \definecolor{brilliantrose}{rgb}{1.0, 0.33, 0.64}

    \begin{groupplot}[
      group style={group size= 2 by 2, horizontal sep=2cm, vertical sep=1.8cm},
      height={6cm},
      width=.47\linewidth,
      ]
    \input{figures/teaser_computation/mislabel_support_gpu}
    \input{figures/teaser_computation/mislabel_support_storage}
    \input{figures/teaser_computation/removal_support_gpu}
    \input{figures/teaser_computation/removal_support_storage}
    \end{groupplot}
    \end{tikzpicture}
\label{fig:compute-efficiency}
\caption{\textbf{Our proposed baseline approach for data attribution achieves high performance while improving computational efficiency and storage requirements.} The wall clock time is computed on an RTX A6000 GPU and memory requirements is computed in GBs (see \cref{app:compute-description}). Mislabel and Removal Support AUC are used to measure the method's accuracy to make counterfactual predictions (see details in \cref{sec:eval-strategy}).}

\end{figure}

% Compute Time A6000
% Moco Resnet-18 800 epochs: 787.5 mins
% Moco Resnet-9 800 epochs: 502.5 mins
% TRAK - 10: 43.29 min
% TRAK - 20: 89.37 min
% TRAK - 100: 447.6 min
% DataModel - 300K: 140,500 min
% DataModel - 50K: 23416.67 min
% DataModel - 10K: 4683.33 min

% Storage Requirements
% Ours - 0.595 GB
% TRAK 10 - 43 GB
% TRAK 20 - 85.1 GB
% TRAK 100 - 423 GB
% DataModels 300K - 2578 GB
% DataModels 50K - 429 GB
% DataModels 10K - 86GB

% Performance Storage and Compute
% AUC Removal + AUC Mislabel on Support

% \micah{should we instead say the brute-force strategy where you enumerate every subset of the training dataset and train on each individually (not just leave 1 out)?} 
In principle, data attribution can be done perfectly by a brute-force leave-$k$-out strategy; simply train the model from scratch many times, removing $k$ data points each time.  The user can then examine the impact of each data point by examining how the corresponding ablated model differs from the original.  Clearly, this procedure is intractable for any realistic problem as there are innumerable subsets, and training even a single machine learning model can be almost prohibitively expensive.  The goal of data attribution research therefore is to approximate this gold standard metric as closely as possible while simultaneously using as little computation as possible.  As such, the field of data attribution is all about trade-offs between accuracy, runtime, and memory.

Existing data attribution approaches gain insights into model behaviors by scraping information from the learning algorithm, such as logits \citep{ilyas22datamodels} or gradients \citep{koh2017understanding, park2023trak}.  Despite this, these techniques still require re-training multiple models on different data subsets, or other compute and memory intensive strategies for better efficacy \citep{ilyas22datamodels, feldman2020neural, koh2017understanding, park2023trak}. Current data attribution approaches quickly become intractable as datasets become larger \citep{basu2021influence, park2023trak} and applications become more realistic, such as attribution for LLMs \citep{grosse_studying_2023}.

In this work, we present a simple approach that outperforms the current state of the art in terms of compute-accuracy trade-offs, and often in terms of raw performance numbers as well.
Given a test image, we use the feature space of a single self-supervised model to retrieve similar images, revealing a compelling association between data attribution and \emph{visual similarity}.
In contrast to existing methods that involve unwieldy model ensembles and extensive computation, our approach shifts the spotlight directly onto the data. Building on prior research, we focus on counterfactual prediction \citep{ilyas22datamodels, park2023trak} for evaluating data attribution techniques.  Based on the intuition that data inherently shapes model behavior, our method does not use any information about the model training process, and yet still rivals the performance of state-of-the-art approaches that do, while using a tiny fraction of the computational resources. Our work shows that contrary to previous work \citep{ilyas22datamodels, park2023trak}, feature representations can serve as a robust baseline for data attribution methods. Our code is available at \url{https://github.com/vasusingla/simple-data-attribution}

% \section{Motivation}
% \label{sec:motivation}
% Focus more on motivation, in the intro. what is needed out of data attribution? Under infinite compute it's easy better compute approaches is the focus.
% Motivation, have more discussion about information being used by the datamodel vs us. We're much more model-agnostic.

\section{Problem Setting}
\label{sec:methodology}

We first define our notation and then discuss evaluation criteria used for data attribution approaches. We borrow notation and evaluation criteria from \citet{ilyas22datamodels} and \citet{park2023trak}. 

\textbf{Notation}: Let $S = \{z_1, z_2, \dots z_n \}$ denote a set of training samples. Each sample $z_i \in S$ represents $z_i = (x_i, y_i)$, where $x_i$ signifies the input image and $y_i$ represents the associated ground truth label. We use $z_t$ to denote an arbitrary evaluation sample not present in the training set. 
% For a model trained on $S$ with parameters $\theta$, we define its correct-class margin on $z$ as $f_\theta(z, S) \in \mathbb{R}$.  
% \tom{This needs to be rigorously defined}

% \micah{not trying to mess with this if it's standard notation, but it seems that a data attribution method can instead be a function into $2^{\mathbb{R}^n}$ which assigns a score to each subset which is not in general recoverable from a function into $\mathbb{R}^n$.}
We denote a data attribution approach as a function $\tau(z, S) \in \mathbb{R}^n$. This function operates on any sample $z$ and a training set $S$, generating a score for each sample within the set $S$. These scores highlight the relative positive or negative impact of individual training samples on the classification of the input sample $z$.

\subsection{Evaluating Attribution Methods}
\label{sec:eval-strategy}
Obtaining ground truth for data attribution has been a challenging problem. Several works have focused on evaluating data attribution methods using alternatives such as Shapley values or leave-one-out influences \citep{koh2017understanding, lundberg2017unified, jia2021scalability}. These approaches however do not scale beyond modest dataset sizes. An alternate line of work evaluates the utility of attribution methods for auxiliary tasks such as active learning or identifying mislabeled or poisoned data samples \citep{liu2021active, jia2021scalability}.

Recent research primarily concentrates on evaluating the performance of data attribution methods through the lens of their capacity to provide accurate counterfactual predictions \citep{park2023trak, ilyas22datamodels}. While these metrics can be computationally demanding, they represent a straightforward, yet valuable, proxy for assessing the effectiveness of attribution approaches.  In our work, we replicate the approach presented in \citet{ilyas22datamodels} and focus on \textbf{data brittleness}. Data brittleness metrics leverage attribution techniques to answer the following question: ``\emph{To what extent are model predictions sensitive to modifications in the training data?}'' Hence, these metrics serve as a means of estimating counterfactual scenarios. To quantify data brittleness, we focus on two distinct types of data support for a validation sample $z_t$. We explain these below:  

\textbf{Data Removal Support:} The smallest subset $R_{r}$, that when removed from the training set $S$, causes an average training run of the model to misclassify $z_t$.

\textbf{Data Mislabel Support:} The smallest training subset $R_{m}$, whose mislabeling causes an average training run of the model to misclassify $z_t$. For each training sample in $R_{m}$, we change the labels to the second-highest predicted class for $z_t$.

 % \vasu{TODO - Explain the naive optimal solution, how this turns data attribution into a better compute problem. }
 % \vasu{TODO - Explain then we perform this over several validation samples and plot the CDF. We also use AUC under CDF for comparing different attribution methods.}
Intuitively, a better data attribution approach should be able to find a smaller subset of training samples that can misclassify $z_t$. We estimate these metrics over a set of validation samples and plot the cumulative distribution (CDF), which represents the probability that a sample's label can be flipped as a function of the data subset size.  In \cref{fig:compute-efficiency}, we compare the Area Under Curve (AUC) of the CDF for the metrics described above across our approach and other attribution methods. 

% We first focus on metrics that characterize data brittleness. 
For a validation sample $z_t$ and a data attribution approach $\tau (z, S)$, we rank the training samples based on decreasing order of positive influence on $z_t$. Then, based on the ranking, we iteratively select and modify a subset of training data. We perform this search, over different subsets to compute the smallest training subset that can cause $z_t$ to be misclassified. Naively, checking all possible subsets would be computationally expensive.  \cite{ilyas22datamodels} check only subsets with certain discrete sizes to keep costs manageable. We instead propose to perform a \textbf{bisection search} to approximate the search for the smallest subset, yielding more accurate results. The bisection search approximation is supported by the observation that several data attribution approaches are additive \citep{park2023trak}. The exact algorithm and details are discussed in \cref{app:computing-support}.

% Based on the above algorithm, we modify the training data and focus on two forms of data brittleness. These are explained below -

 % These approaches measure the efficacy of a method's ability to make accurate counterfactual predictions.

%Linear datamodeling score (LDS) is another related metric used for evaluation of data attribution methods \citep{ilyas22datamodels, park2023trak}. This metric is defined as follows - 

Linear Datamodeling Score (LDS) is another related metric used for the evaluation of data attribution methods \citep{ilyas22datamodels, park2023trak}. Note that the LDS metric focuses on counterfactual predictions for \emph{arbitrary} changes in training data. In contrast, data brittleness serves to quantify the accuracy of counterfactual predictions using \emph{targeted} changes to training data based on a specific validation sample. Thus, the latter metric serves as a better proxy for the data attribution method's usefulness as a debugging tool. In this work, we emphasize performance on data brittleness and provide results for the LDS metric in \cref{app:lds-scores}.

% \tom{I'm on the fence about whether to even talk about linear score if we don't include them.  As I recall you were going to put them in the appendix though?  In which case its worth bringing up.  It could probably just be a sentence though instead of highlighting with a big bolded definition.}

% A related evaluation metric is based on the accuracy of counterfactual predictions for arbitrary changes in training data \cite{ilyas22datamodels, park2023trak}.  However, this metric isn't useful for understanding model biases. This metric is based on the accuracy of \emph{counterfactual predictions} using targeted changes to training data based on a specific validation sample. Thus, it serves as a useful proxy for the usefulness of quantifying model biases and serves as a useful debugging tool \cite{ilyas22datamodels, shah2023modeldiff}.
% \vasu{write note on optimal guidance approach}

\section{Our Approach \& Baselines}
\label{sec:our-approach-and-baselines}

Our approach utilizes the feature space of a neural network to extract features from a validation sample $z_t$ and each training sample in $S$. We then compute the attribution scores by measuring the distance in feature space between $z_t$ and each training sample in $S$. Prior works have tried similar approaches and claimed them to be ineffective for counterfactual estimation \citep{park2023trak, ilyas22datamodels}. In the next sections, we describe the details of our approach and discuss our baselines.
% \vasu{TODO, add a small description about comparing image embeddings in feature space to retrieve top-k nearest neighbors. }

\begin{figure}[!htb]
    \centering
    \begin{tikzpicture}
    \centering
    \definecolor{darkgray141160203}{RGB}{141,160,203}
    \definecolor{dimgray85}{RGB}{85,85,85}
    \definecolor{gainsboro229}{RGB}{229,229,229}
    \definecolor{lightgray204}{RGB}{204,204,204}
    \definecolor{mediumaquamarine102194165}{RGB}{102,194,165}
    \definecolor{orchid231138195}{RGB}{231,138,195}
    \definecolor{salmon25214198}{RGB}{252,141,98}
    \definecolor{amethyst}{rgb}{0.6, 0.4, 0.8}
    \definecolor{bleudefrance}{rgb}{0.19, 0.55, 0.91}
    \definecolor{blush}{rgb}{0.87, 0.36, 0.51}
    \definecolor{brilliantrose}{rgb}{1.0, 0.33, 0.64}
    \begin{groupplot}[
        group style={group size= 1 by 1, horizontal sep=0cm, vertical sep=0cm},
        height={6cm},
        width=.55\linewidth]
    \nextgroupplot[
        xlabel={Number of Training Samples \textbf{Removed}},
        ylabel={Frac. of CIFAR-10 Misclassified},
        legend pos={north east},
        axis background/.style={fill=gainsboro229},
        axis line style={white},
        legend cell align={left},
        legend columns=1,
        legend style={
        column sep=0.05cm,
        fill opacity=0.8,
        draw opacity=1,
        text opacity=1,
        at={(1.05, 0.9)},
        anchor=north west,
        fill=none,
        draw=black,
        },
        ymajorgrids=true,
        mark size=1.0pt,
        xmin=0, xmax=1280,
        xmajorgrids, xminorgrids,
        x grid style={white},
        y grid style={white},
        tick align=outside,
        tick pos=left,
    ]

    % Datamodels
    % \addplot[ultra thick, color=mediumaquamarine102194165] table [x=x, y=y, col sep=comma] {data/brittleness_remove/dmodel.csv};
    % \addlegendentry{Datamodels ($300$k models)}

    % L2 MoCo ResNet-18
    % \addplot[ultra thick, color=orchid231138195, dashed] table [x=x, y=y, col sep=comma] {data/brittleness_remove/L2-MoCo.csv};
    % \addlegendentry{L2 MoCo}

    % L2 MoCo ResNet-9
    \addplot[ultra thick, color=amethyst, dashed] table [x=x, y=y, col sep=comma] {data/brittleness_remove/L2-MoCo-ResNet9.csv};
    \addlegendentry{$\ell_2$ MoCo}
    
    % ESVM MoCo ResNet-9
    \addplot[ultra thick, color=amethyst] table [x=x, y=y, col sep=comma] {data/brittleness_remove/ESVM-MoCo-ResNet9.csv};
    \addlegendentry{ESVM MoCo}

    % ESVM MoCo ResNet-18
    \addplot[ultra thick, color=bleudefrance] table [x=x, y=y, col sep=comma] {data/brittleness_remove/ESVM-MoCo.csv};
    \addlegendentry{ESVM MoCo (ResNet-18)}    

    % L2 ResNet-9
    \addplot[ultra thick, color=salmon25214198, dashed] table [x=x, y=y, col sep=comma] {data/brittleness_remove/L2-ResNet9.csv};
    \addlegendentry{$\ell_2$ Supervised}

    % ESVM ResNet-9
    \addplot[ultra thick, color=salmon25214198] table [x=x, y=y, col sep=comma] {data/brittleness_remove/ESVM-ResNet9.csv};
    \addlegendentry{ESVM Supervised}

    % \addplot[ultra thick, color=black]{}
    \end{groupplot}
\end{tikzpicture}
    \caption{\textbf{Self-supervised features are more effective than supervised and are best compared using an ESVM}. Self-supervised features from MoCo can be used to find smaller data support than standard supervised features. For a larger fraction of test samples, ESVM distance is more effective than $\ell_2$ distance at ranking train images to select smaller data removal support.}
\label{fig:k-leave-out-l2-vs-esvm}
\end{figure}
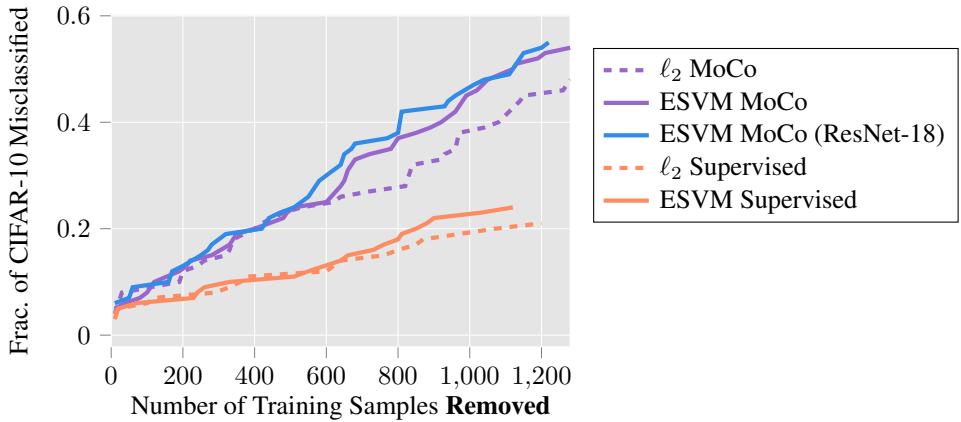

\subsection{Design Choices}
\label{sec:design-choices}

Our data attribution approach relies on the comparison of image embeddings, and in doing so, we make decisions regarding the choice of feature extractor, the subset of training images to compare, and the distance function.

\textbf{Feature extractor.} We find that the learning paradigm used to train a feature extractor heavily influences the estimation of data support. For example, embeddings from a ResNet-9 trained using a self-supervised learning objective (MoCo, \citep{he2020momentum}) can be used to find smaller support sets than the same model trained in a supervised manner (See $\ell_2$ MoCo vs $\ell_2$ Supervised in \cref{fig:k-leave-out-l2-vs-esvm}). With the exception of DINO \citep{caron2021emerging}, all self-supervised feature extractors perform better than their supervised counterpart (see \cref{app:additional-self-supervised} \cref{fig:k-leave-out-l2-vs-esvm-appendix}). We found that MoCo features outperform other self-supervised approaches in both data removal support and mislabeling support scenarios, leading us to select a MoCo model as our preferred feature extractor. We find that the ResNet-18 backbone provides better support estimates than ResNet-9, and hence use it as default for all our experiments.

% Labeling Function
% Discuss datamodels and trak. They tried without labels and it didn't work.
\textbf{Subset of train images.} In \cref{app:chosen-subset-of-train} \cref{fig:k-flip-l2-all-vs-same-class-appendix}, we show that choosing a support set from training images of class the same class $y$ as the target $z_t = (x, y)$ is critical, \ie given a target image of an airplane, we only rank airplane training images. 

\textbf{Distance function.} When measuring the distance between two embeddings, Euclidean distance ($\ell_2$) is a common choice \citep{ilyas22datamodels, park2023trak}. Cosine distance and Mahalanobis distance have also been used to measure similarity, but these were found to perform similarly to Euclidean distances in previous work \citep{hanawa2021evaluation, ilyas22datamodels, park2023trak}.

However, we find that measuring distance as distance to the hyperplane of an Exemplar SVM (ESVM) improves image similarity \citep{malisiewicz11exemplar}. To compute this metric, we train a linear SVM using one positive sample (the target embedding) and treat all other samples (the remaining embeddings of the same class) as negative samples. In this way, the decision boundary, and consequently the distance function, is defined largely by unique dimensions of the target with respect to all embeddings of the same class. In \cref{fig:k-leave-out-l2-vs-esvm}, we demonstrate how using distance to the hyperplane of an ESVM yields better removal support estimates than $\ell_2$ distance.

\subsection{Baselines}

% We focus on Datamodels \citep{ilyas22datamodels} and TRAK \citep{park2023trak} as these data attribution approaches show state-of-the-art results for counterfactual estimation. While other attribution approaches can be applied to 

\textbf{Datamodels} \citep{ilyas22datamodels}: In the \textit{Datamodeling} framework, the end-to-end training and evaluation of deep neural networks is approximated with a parametric function. Surprisingly, optimizing a linear function is enough to predict model outputs reasonably well, when given a training data subset. By collecting a large dataset of subset-output pairs, \cite{ilyas22datamodels} demonstrate that such a linear mapping can accurately predict the correct-class margin. Among other use-cases, these Datamodels are shown to be effective at counterfactual predictions and identifying visually similar train-test samples. But Datamodeling is prohibitively expensive, requiring the training of hundreds of thousands of models (300,000 in the original work) to generate optimal subset-output data. Unfortunately, this limitation makes Datamodeling intractable for all but small toy problems.

\textbf{TRAK} \citep{park2023trak}: By approximating models with a kernel machine, \textit{Tracing with the Randomly-projected After Kernels} (TRAK) makes progress toward reducing the computational cost of data attribution by reducing dimensionality with random projections and ensembling over independently trained models. However, the method tends to only work well with more than a dozen model checkpoints and a large projection dimension for the model gradients, the storage of which can surpass 80GB when using a ResNet-9 on CIFAR-10. Compared to Datamodels, TRAK gains in runtime are paid for in storage space.

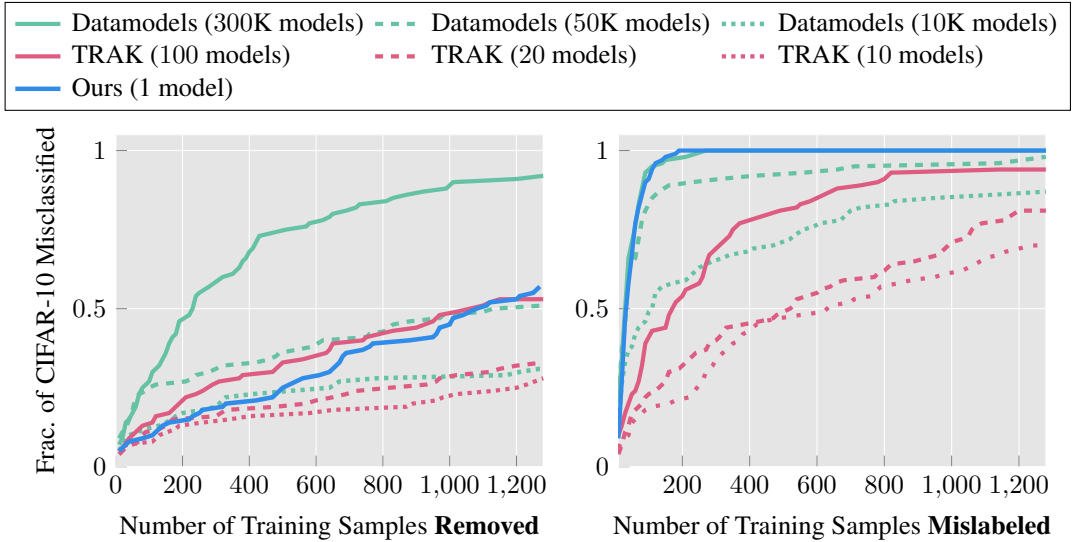
\begin{figure}[tb]
\centering

\begin{tikzpicture}
    \centering
    \definecolor{darkgray141160203}{RGB}{141,160,203}
    \definecolor{dimgray85}{RGB}{85,85,85}
    \definecolor{gainsboro229}{RGB}{229,229,229}
    \definecolor{lightgray204}{RGB}{204,204,204}
    \definecolor{mediumaquamarine102194165}{RGB}{102,194,165}
    \definecolor{orchid231138195}{RGB}{231,138,195}
    \definecolor{salmon25214198}{RGB}{252,141,98}
    \definecolor{amethyst}{rgb}{0.6, 0.4, 0.8}
    \definecolor{bleudefrance}{rgb}{0.19, 0.55, 0.91}
    \definecolor{blush}{rgb}{0.87, 0.36, 0.51}
    \definecolor{brilliantrose}{rgb}{1.0, 0.33, 0.64}

    \begin{groupplot}[
        group style={group size= 2 by 1, horizontal sep=1.cm, vertical sep=0cm},
        height={6cm},
        width=.52\linewidth]
    \nextgroupplot[
        xlabel={Number of Training Samples \textbf{Removed}},
        ylabel={Frac. of CIFAR-10 Misclassified},
        ylabel near ticks, 
        legend pos={north east},
        axis background/.style={fill=gainsboro229},
        axis line style={white},
        legend cell align={left},
        legend columns=3,
        legend style={
        % row sep=0.1cm,
        column sep=0.05cm,
        fill opacity=0.8,
        draw opacity=1,
        text opacity=1,
        at={(-0.26, 1.4)},
        anchor=north west,
        % draw=gainsboro229,
        % fill=gainsboro229,
        fill=none,
        draw=black,
        },
        ymajorgrids=true,
        mark size=1.5pt,
        xmin=0, xmax=1280,
        ymin=0, ymax=1.05,
        xmajorgrids, xminorgrids,
        x grid style={white},
        y grid style={white},
        tick pos=left,
        % no marks,
        every axis plot/.append style={ultra thick}
    ]

    % \addplot[ultra thick, color=mediumaquamarine102194165] table [x=x, y=y, col sep=comma] {data/brittleness_methods/ESVM-BYOL.csv};
    % \addlegendentry{BYOL-ESVM}
    \addplot[color=mediumaquamarine102194165] table [x=x, y=y, col sep=comma] {data/brittleness_remove/dmodel.csv};
    \addlegendentry{Datamodels ($300$K models)}

    \addplot[color=mediumaquamarine102194165, dashed] table [x=x, y=y, col sep=comma] {data/brittleness_remove/dmodel-50kmodels.csv};
    \addlegendentry{Datamodels ($50$K models)}

    \addplot[color=mediumaquamarine102194165, dotted, line width=0.6mm] table [x=x, y=y, col sep=comma] {data/brittleness_remove/dmodel-10kmodels.csv};
    \addlegendentry{Datamodels ($10$K models)}
    
    \addplot[color=blush] table [x=x, y=y, col sep=comma] {data/brittleness_remove/traker100models20480proj.csv};
    \addlegendentry{TRAK ($100$ models)}

    \addplot[color=blush, dashed]table [x=x, y=y, col sep=comma] {data/brittleness_remove/traker20models20480proj.csv};
    \addlegendentry{TRAK ($20$ models)}

    \addplot[color=blush, dotted]table [x=x, y=y, col sep=comma] {data/brittleness_remove/traker10models20480proj.csv};
    \addlegendentry{TRAK ($10$ models)}

    \addplot[color=bleudefrance, line width=0.6mm] table [x=x, y=y, col sep=comma] {data/brittleness_remove/MOCOESVM.csv};
    \addlegendentry{Ours (1 model)}

    % \addplot[ultra thick, color=mediumaquamarine102194165] table [x=x, y=y, col sep=comma] {data/brittleness_methods/Datamodel50k.csv};
    % \addlegendentry{Datamodels ($50$k models)}

    % \addplot[ultra thick, color=salmon25214198 ] table [x=x, y=y, col sep=comma] {data/brittleness_methods/RepresentationSimilarity.csv};
    % \addlegendentry{Representation Sim. ($100$ models)}

    % Plot 2
    \nextgroupplot[
        xlabel={Number of Training Samples \textbf{Mislabeled}},
        axis background/.style={fill=gainsboro229},
        axis line style={white},
        legend cell align={left},
        ymajorgrids=true,
        mark size=1.0pt,
        xmin=10, xmax=1280,
        ymin=0, ymax=1.05,
        xmajorgrids, xminorgrids,
        x grid style={white},
        y grid style={white},
        tick pos=left,
        no marks,
        every axis plot/.append style={ultra thick}
    ]

    % \addplot[ultra thick, color=mediumaquamarine102194165] table [x=x, y=y, col sep=comma] {data/brittleness_methods/ESVM-BYOL.csv};
    % \addlegendentry{BYOL-ESVM}
    \addplot[color=mediumaquamarine102194165] table [x=x, y=y, col sep=comma] {data/brittleness_flip/dmodel.csv};
    % \addlegendentry{Datamodels ($300$k models)}

    \addplot[color=mediumaquamarine102194165, mark=*, dashed] table [x=x, y=y, col sep=comma] {data/brittleness_flip/dmodel-50kmodels.csv};
    % \addlegendentry{Datamodels ($50$k models)}

    \addplot[color=mediumaquamarine102194165, mark=*, dotted] table [x=x, y=y, col sep=comma] {data/brittleness_flip/dmodel-10kmodels.csv};
    % \addlegendentry{Datamodels ($10$k models)}
    
    \addplot[color=blush, mark=o] table [x=x, y=y, col sep=comma] {data/brittleness_flip/traker100models20480proj.csv};
    % \addlegendentry{TRAK ($100$ models)}

    \addplot[color=blush, mark=o, dashed]table [x=x, y=y, col sep=comma] {data/brittleness_flip/traker20models20480proj.csv};
    % \addlegendentry{TRAK ($20$ models)}

    \addplot[color=blush, mark=o, loosely dotted]table [x=x, y=y, col sep=comma] {data/brittleness_flip/traker10models20480proj.csv};
    % \addlegendentry{TRAK ($10$ models)}
    
    \addplot[color=bleudefrance, mark=square*] table [x=x, y=y, col sep=comma] {data/brittleness_flip/MOCOESVM.csv};
    % \addlegendentry{Ours (1 model)}

    % \addplot[ultra thick, color=mediumaquamarine102194165] table [x=x, y=y, col sep=comma] {data/brittleness_methods/Datamodel50k.csv};
    % \addlegendentry{Datamodels ($50$k models)}

    % \addplot[ultra thick, color=salmon25214198 ] table [x=x, y=y, col sep=comma] {data/brittleness_methods/RepresentationSimilarity.csv};
    % \addlegendentry{Representation Sim. ($100$ models)}

    \end{groupplot}
\end{tikzpicture}

\caption{\textbf{Our baseline approach uses only a single model and outperforms TRAK and Datamodels using 20 and 10,000 models for data brittleness metrics.}  We estimate data removal and data mislabel support for 100 random CIFAR-10 test samples using a ResNet-9 model and plot the cumulative distribution using our approach and other baselines. The number of models used by each approach is also shown.  For data removal support, using only a single model our proposed approach outperforms  TRAK \citep{park2023trak} using 20 models and Datamodels \citep{ilyas22datamodels} using 10,000 models. For data mislabel support, we outperform TRAK using 100 models and perform equivalent to Datamodels using 300,000 models.}
\label{fig:k-leave-out}
\end{figure}

% legend bg color
% figure bg color
% maybe thickness to contrast models, and or linestyle
% color of Ours 
% gridline black/gray

\section{Counterfactual Estimation}
\label{sec:data-britleness}
We evaluate these approaches and our proposed baseline data attribution for a number of classification examples in computer vision, focusing on datasets such as CIFAR-10 and ImageNet, which are small enough to allow for some comparison with the more expensive approaches of TRAK and Datamodels.

\subsection{Experimental Setup}
\label{sec:experimental-setup}

\textbf{Training Setup:} We estimate the approximate data removal and data mislabel support for CIFAR-10 and ImageNet. As computing the data support for even a single validation sample requires training multiple models, we restrict ourselves to a reasonably small set of validation samples. We use the same validation samples across all attribution methods. To accelerate the training of these models, we use the FFCV library \citep{leclerc2023ffcv}. 

For CIFAR-10 \citep{krizhevsky2009learning}, we train ResNet-9 \footnote{https://github.com/wbaek/torchskeleton/blob/master/bin/dawnbench/cifar10.py} and MobileNetV2 \citep{sandler2018mobilenetv2}  models for 24 epochs using a batch size of 512, momentum of 0.9, label smoothing of 0.1,  with a cyclic learning schedule, with a maximum value of 0.5. The test accuracy for these models without any modification to training data is above $92\%$. We randomly selected 100 validation samples, in a class-balanced manner for our brittleness metrics.  We remove or mislabel a maximum of $1280$ training samples for each validation sample. Our training setup is similar to \cite{ilyas22datamodels}.

For ImageNet \citep{deng2009imagenet}, we train ResNet-18 \citep{He2015ResNet} models for 16 epochs, using a batch size of 1024. We train on 160$\times$160 resolution images for the first 11 epochs and increase the training resolution to 192$\times$192 for the last 5 epochs. The other hyperparameters are kept the same as CIFAR-10. These models achieve a top-1 validation accuracy of $67\%$. We randomly selected 30 validation samples, from a subset of validation samples that are not misclassified by 4 ResNet-18 models on average. We removed or mislabeled a maximum of $1000$ training samples for each validation sample.

\textbf{Baselines and Our Setup:} To estimate TRAK scores on CIFAR-10, we train 100 ResNet-9 models and use a projection dimension of $20480$. To estimate scores on ImageNet, we train 4 ResNet-18 models and use a projection dimension of $4096$. Computing TRAK scores using 4 models already requires $160$ GB of storage space, hence we refrain from using a larger ensemble of models.

For Datamodels, we download the pre-trained weights optimized using outputs from 300K ResNet-9 models with $50\%$ random subsets.\footnote{https://github.com/MadryLab/datamodels-data} We also download the binary masks and margins to train our own Datamodels on outputs from 10K and 50K ResNet-9 models, using another 10K models for validation. Since Datamodels are extremely compute-intensive and require training hundreds of thousands of models, we cannot include them as a baseline on ImageNet. 

For our baseline approach to train self-supervised models, we use the Lightly library \citep{susmelj2020lightly}. We train a ResNet-18 model using MoCo \citep{he2020momentum} for 800 epochs on CIFAR-10, using the Lightly benchmark code.\footnote{https://docs.lightly.ai/self-supervised-learning/getting\_started/benchmarks.html} On ImageNet, we download a pre-trained ResNet-50 model trained using MoCo.\footnote{https://github.com/facebookresearch/moco} For our approach, we always use a single model. We denote Datamodels using N models as Datamodels (N), and similarly for TRAK. 

% We use these models to extract

\subsection{CIFAR-10 Data Brittleness}

% Remove Case
% traker100models20480proj: 22, 34
% traker10models20480proj: 47, 4
% traker20models20480proj: 41, 11
% dmodel_10kmodels: 37, 14
% dmodel_50kmodels: 16, 40
% dmodel: 0, 86

% Mislabel Case

% traker100models20480proj: 79, 14
% dmodel: 32, 49
% traker10models20480proj: 90, 3
% traker20models20480proj: 89, 2
% dmodel_10kmodels: 68, 19
% dmodel_50kmodels: 35, 44

% \begin{table}[tb]
% \centering
% \begin{tabular}{cc|cc|cc}
% \toprule
% \multicolumn{1}{l}{} & \multirow{2}{*}{Models Used} & \multicolumn{2}{c|}{Mislabel Support} & \multicolumn{2}{c}{Removal Support} \\
% \multicolumn{1}{l}{} &  & Ours \textless Other & Ours = Other & Ours \textless Other & Ours = Other \\
% \midrule
% \multirow{3}{*}{DataModels} & 300,000 & 32 & 19 & 0 & 14 \\
%  & 50,000 & 35 & 21 & 16 & 44 \\
%  & 10,000 & 68 & 13 & 37 & 49 \\
%  \midrule
% \multirow{3}{*}{TRAK}  & 100 & 79 & 7 & 22 & 44 \\
%  & 20 & 89 & 9 & 41 & 48 \\
% & 10 & 90 & 7 & 47 & 49 \\
%  \bottomrule
% \end{tabular}

% \end{table}

\begin{figure}[tb]
    \centering
    \resizebox{\textwidth}{!}{
    \includegraphics[]{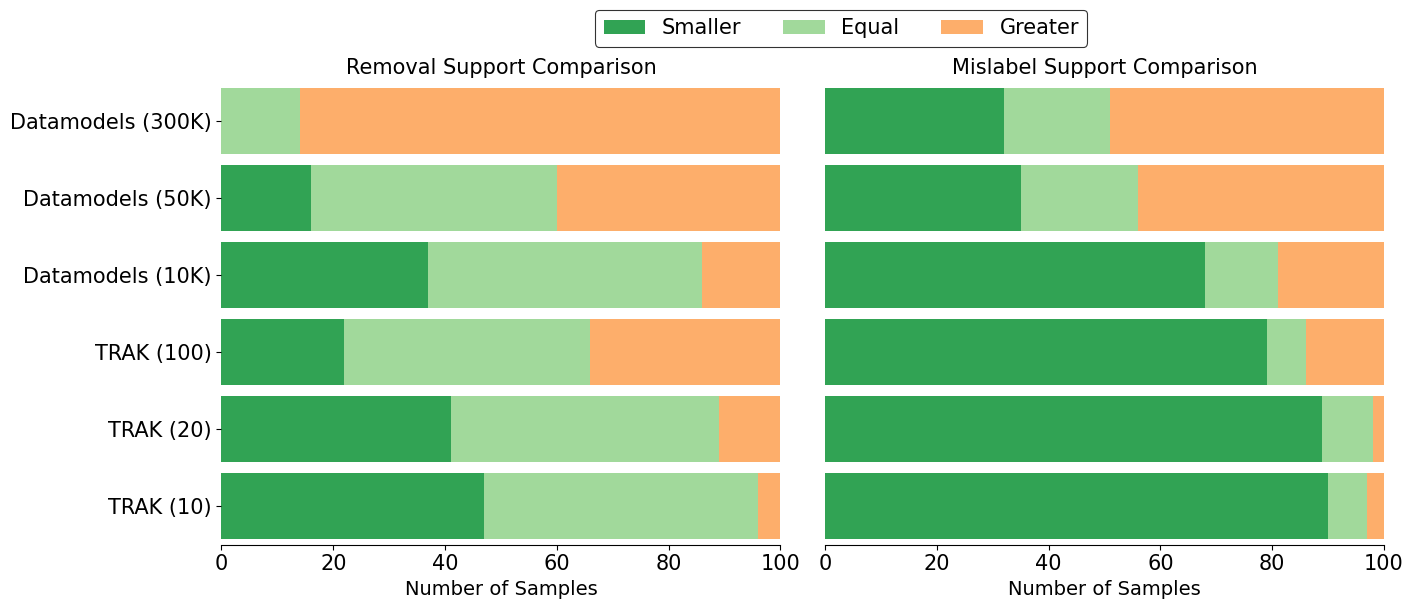}
    }
    \caption{Compared to instances of Datamodels and TRAK, we check whether our data support estimates are smaller, equal, or larger for all 100 validation samples. For 32 samples, our proposed method can find smaller data mislabel support compared to Datamodels (300k models). Even for the data removal case, our approach can find an equivalent support estimate to Datamodels (300k models) for 14 samples.}
    \label{tab:win-rate}
\end{figure}

 In \cref{fig:k-leave-out}, we present the distribution of estimated data removal values for CIFAR-10. Our findings reveal that employing a single model with a MoCo backbone \citep{he2020momentum} for data removal support proves more effective than employing Datamodels with 10,000 models and TRAK with 20 models.  Our approach and Datamodels (10K) identify that 23\% samples can be misclassified by removing fewer than 500 (example-specific) training samples while TRAK (20) can only identify 16\%. For support sizes up to 1280 images, our approach identifies 55\% of validation samples,  whereas TRAK (20) and Datamodels (10K) can only identify 28\% and 31\% samples respectively.

In the same figure, we also depict the distribution of estimated data mislabel support for CIFAR-10. Here, our approach outperforms TRAK (100) and approaches the performance of Datamodels (300K). Here, our approach identifies 47\% of CIFAR-10 validation samples that can be misclassified by mislabeling less than 30 training samples! In contrast, TRAK (100) performs poorly identifying only 20\% of these samples. DataModels (300K) can identify 50\% of validation samples marginally surpassing our performance. 
% Using our method, we uncovered 67 and 91 out of 100 validation samples with data mislabel support of fewer than 50 and 100 samples, respectively. By contrast, TRAK (100) can only detect 23 and 41 samples with the corresponding support sizes. Datamodels (300K) enable us to find 71 and 93 samples with data mislabel support below 50 and 100 samples, respectively, marginally surpassing our estimates.
 
In \cref{tab:win-rate}, we further inspect how well our baseline approach works for each validation sample. We compare the individual estimated support sizes for all 100 samples using our approach versus other baselines. Our results show that for data removal support, across 16\% of validation samples, our estimated data removal support is smaller than those of Datamodels (50K). For 44\% of the samples our data removal estimates match TRAK and Datamodels (50K). For data mislabel support, our approach finds a smaller support estimate than Datamodels and TRAK for 32\% and 79\% of the validation samples.
% and 22 out of 100 validation samples, TRAK (100) 
 
 While our baseline approach cannot outperform Datamodels (300K) on data removal, our performance on the data mislabel support is nearly the same. Our baseline approach of using a single self-supervised model can thus serve as a simple, compute, and storage-efficient alternative to estimate data brittleness. 
 
\subsection{ImageNet Data Brittleness}

\begin{figure}
    \centering
    \begin{tikzpicture}
    \centering
    \definecolor{darkgray141160203}{RGB}{141,160,203}
    \definecolor{dimgray85}{RGB}{85,85,85}
    \definecolor{gainsboro229}{RGB}{229,229,229}
    \definecolor{lightgray204}{RGB}{204,204,204}
    \definecolor{mediumaquamarine102194165}{RGB}{102,194,165}
    \definecolor{orchid231138195}{RGB}{231,138,195}
    \definecolor{salmon25214198}{RGB}{252,141,98}
    \definecolor{amethyst}{rgb}{0.6, 0.4, 0.8}
    \definecolor{bleudefrance}{rgb}{0.19, 0.55, 0.91}
    \definecolor{blush}{rgb}{0.87, 0.36, 0.51}
    \definecolor{brilliantrose}{rgb}{1.0, 0.33, 0.64}

    \begin{groupplot}[
        group style={group size= 1 by 1, horizontal sep=1.cm, vertical sep=0cm},
        height={6cm},
        width=.55\linewidth]
    \nextgroupplot[
        xlabel={Number of Training Samples \textbf{Removed}},
        % xlabel style={
        % font=\small,
        % },
        ylabel={Frac. of ImageNet Misclassified},
        % ylabel style={
        % font=\small,
        % },
        ylabel near ticks, 
        legend pos={north east},
        axis background/.style={fill=gainsboro229},
        axis line style={white},
        legend cell align={left},
        legend columns=1,
        legend style={
        % font=\small,
        % row sep=0.1cm,
        column sep=0.05cm,
        fill opacity=0.8,
        draw opacity=1,
        text opacity=1,
        at={(1.05, 0.9)},
        anchor=north west,
        fill=none,
        draw=black,
        },
        ymajorgrids=true,
        mark size=1.5pt,
        xmin=0, xmax=1000,
        ymin=0, ymax=1.05,
        xmajorgrids, xminorgrids,
        x grid style={white},
        y grid style={white},
        tick pos=left,
        % no marks,
        every axis plot/.append style={ultra thick}
    ]
    
    \addplot[color=blush] table [x=x, y=y, col sep=comma] {data/brittleness_remove_imagenet/traker_4.csv};
    \addlegendentry{TRAK ($4$ models)}

    \addplot[color=blush, dashed]table [x=x, y=y, col sep=comma] {data/brittleness_remove_imagenet/traker_1.csv};
    \addlegendentry{TRAK ($1$ model)}

    \addplot[color=bleudefrance, line width=0.6mm] table [x=x, y=y, col sep=comma] {data/brittleness_remove_imagenet/moco_esvm.csv};
    \addlegendentry{Ours (1 model)}
    
\end{groupplot}
\end{tikzpicture}
% \label{fig:k-leave-out-imagenet}
% \caption{Data Removal Support on ImageNet}
    \caption{\textbf{Our method yields better upper bounds on support size compared to TRAK-4, which requires more storage than the ImageNet dataset itself.} We estimate data removal support for 30 random ImageNet validation samples and plot the CDF of estimates.}
    \label{fig:k-leave-out-imagenet}
\end{figure}
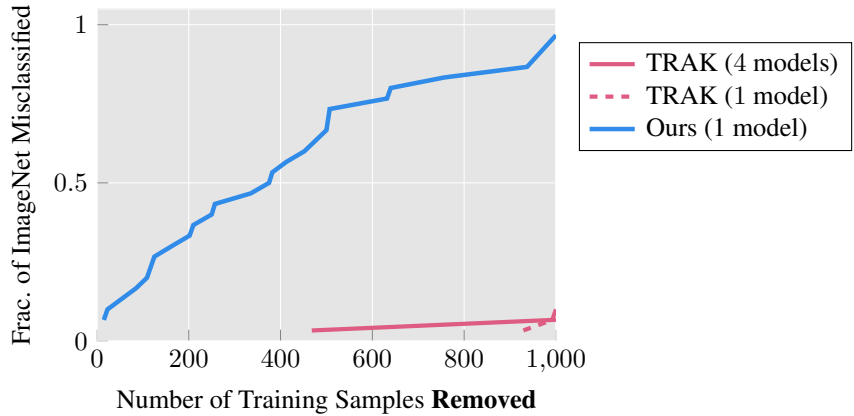

 In \cref{fig:k-leave-out-imagenet}, we show our results for data removal on ImageNet. Our results show that for 4 and 16 of the 30 validation samples our estimated data removal support is less than 16 and 130 training samples respectively. In contrast, TRAK (1) and TRAK (4) do not scale well to ImageNet at all and provide much looser data removal estimates. \uline{We again emphasize that even scaling to TRAK with 10 models would require around 400 GB of storage space, by our estimate.} This highlights the scalability of our baseline approach where a single self-supervised MoCo backbone can provide more accurate data removal estimates than other existing data attribution methods.

\begin{figure}[!htb]
\centering

\begin{tikzpicture}
    \centering
    \definecolor{darkgray141160203}{RGB}{141,160,203}
    \definecolor{dimgray85}{RGB}{85,85,85}
    \definecolor{gainsboro229}{RGB}{229,229,229}
    \definecolor{lightgray204}{RGB}{204,204,204}
    \definecolor{mediumaquamarine102194165}{RGB}{102,194,165}
    \definecolor{orchid231138195}{RGB}{231,138,195}
    \definecolor{salmon25214198}{RGB}{252,141,98}
    \definecolor{amethyst}{rgb}{0.6, 0.4, 0.8}
    \definecolor{bleudefrance}{rgb}{0.19, 0.55, 0.91}
    \definecolor{blush}{rgb}{0.87, 0.36, 0.51}
    \definecolor{brilliantrose}{rgb}{1.0, 0.33, 0.64}
    \begin{groupplot}[
        group style={group size= 1 by 1, horizontal sep=0cm, vertical sep=0cm},
        height={6cm},
        width=.55\linewidth]
    \nextgroupplot[
        xlabel={Number of Training Samples \textbf{Removed}},
        ylabel={Frac. of CIFAR-10 Misclassified},
        legend pos={north east},
        axis background/.style={fill=gainsboro229},
        axis line style={white},
        legend cell align={left},
        legend columns=1,
        legend style={
        column sep=0.05cm,
        fill opacity=0.8,
        draw opacity=1,
        text opacity=1,
        at={(1.05, 0.9)},
        anchor=north west,
        fill=none,
        draw=black,
        },
        ymajorgrids=true,
        mark size=1.0pt,
        xmin=0, xmax=1280,
        xmajorgrids, xminorgrids,
        x grid style={white},
        y grid style={white},
        tick align=outside,
        tick pos=left,
    ]

    \addplot[color=mediumaquamarine102194165, line width=0.6mm] table [x=x, y=y, col sep=comma] {data/brittleness_remove_mobilenetv2/dmodel.csv};
    \addlegendentry{Datamodels ($300$K models)}

    \addplot[color=mediumaquamarine102194165, dashed, line width=0.6mm] table [x=x, y=y, col sep=comma] {data/brittleness_remove_mobilenetv2/dmodel_50kmodels.csv};
    \addlegendentry{Datamodels ($50$K models)}

    \addplot[color=mediumaquamarine102194165, dotted, line width=0.6mm] table [x=x, y=y, col sep=comma] {data/brittleness_remove_mobilenetv2/dmodel_10kmodels.csv};
    \addlegendentry{Datamodels ($10$K models)}
    
    \addplot[color=blush, line width=0.6mm] table [x=x, y=y, col sep=comma] {data/brittleness_remove_mobilenetv2/traker100models20480proj.csv};
    \addlegendentry{TRAK ($100$ models)}

    \addplot[color=blush, dashed, line width=0.6mm]table [x=x, y=y, col sep=comma] {data/brittleness_remove_mobilenetv2/traker20models20480proj.csv};
    \addlegendentry{TRAK ($20$ models)}

    \addplot[color=blush, dotted, line width=0.6mm]table [x=x, y=y, col sep=comma] {data/brittleness_remove_mobilenetv2/traker10models.csv};
    \addlegendentry{TRAK ($10$ models)}

    \addplot[color=bleudefrance, line width=0.6mm] table [x=x, y=y, col sep=comma] {data/brittleness_remove_mobilenetv2/Moco_800epoch_esvm_c01_only_req_test_idx.csv};
    \addlegendentry{Ours (1 model)}
    
    \end{groupplot}
\end{tikzpicture}
\caption{\textbf{Our baseline approach is model agnostic and performs well across different architectures.} We evaluate how attribution scores transfer from one architecture transfer to another. We use ResNet-9 scores for TRAK and DataModels and estimate data removal support for MobileNetV2. For our approach, we use the same ResNet-18 backbone.}
\label{fig:k-leave-out-mobilenetv2}
\end{figure}
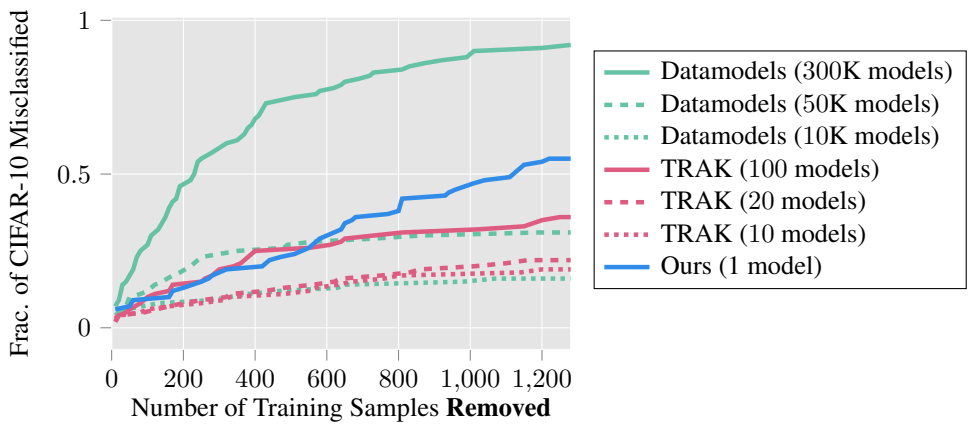
\subsection{Transfer to different architecture}

Datamodels and TRAK utilize information tied to the model architecture such as gradients or logits from an ensemble of models. However, different neural network architectures are known to exploit similar biases and output similar predictions \citep{mania2019model, toneva2018empirical}. In order to better understand how data may be shaping these biases we test how well attribution scores from these approaches transfer to other architectures. Since our approach does not use any information about the model architecture and only leverages the data, we expect our baseline approach to transfer across different architectures. 
  %Thus, we expect attribution scores from one architecture to transfer well to a different architecture. 

% A key aspect to highlight is that Datamodels and TRAK utilize information tied to the model architecture. In contrast, our approach

In \cref{fig:k-leave-out-mobilenetv2}, we compare TRAK, Datamodels, and our attribution scores and evaluate them on a MobileNetV2 architecture \citep{sandler2018mobilenetv2}. The results show that our approach using ResNet-18 continues to predict accurate data removal estimates surpassing TRAK (100) and Datamodels (50K), which suffer a large degradation in performance. Datamodels (300K) also suffer degradation in performance but provide tighter estimates than our approach. This suggests that while simply relying on visual similarity may be useful for efficiently predicting counterfactuals, additional biases within the architecture may also have an influence.

% \subsection{Computational Cost}

% \vasu{TODO - Add a blurb that our approach scales better, as TRAK and Datamodels require more compute as model architecture size, while our setup is more agnostic}

% Actual support comparisons for 100 samples - 
% Number of times Moco < TRAK (100): 22
% Number of times Moco < DataModel: 0
% Number of times Moco > TRAK (100): 34
% Number of times Moco > DataModel: 86

% AUC numbers
% DataModels (300K): 0.7084374999999998
% MOCO ESVM: 0.31363281249999997
% TRAK (10): 0.17523437500000003
% TRAK (20): 0.21644531250000001
% TRAK (100): 0.35527343750000007

% Actual support comparisons for 100 samples - 
% Number of times Moco < TRAK (100): 79
% Number of times Moco < DataModel: 32
% Number of times Moco > TRAK (100): 14
% Number of times Moco > DataModel: 49

% AUC numbers
% DataModel (300K): 0.966796875
% MOCO ESVM: 0.965390625
% TRAK (10): 0.46953124999999996
% TRAK (20): 0.5369531249999999
% TRAK (100): 0.7750390625000001

% Histogram for comparison of support sizes for datamodels and us?
% Win - rate Table?

\section{Discussion}
\label{sec:discussion}
\subsection{Role of Visual Similarity}

% Describe visual similarity is a good enough metric for counterfactual 
% DataModels and TRAK claimed different, but they 
% L2 vs ESVM in the appendix

\begin{figure}[t]
    \centering
    \resizebox{\textwidth}{!}{
    \centerline{\includegraphics[width=\textwidth]{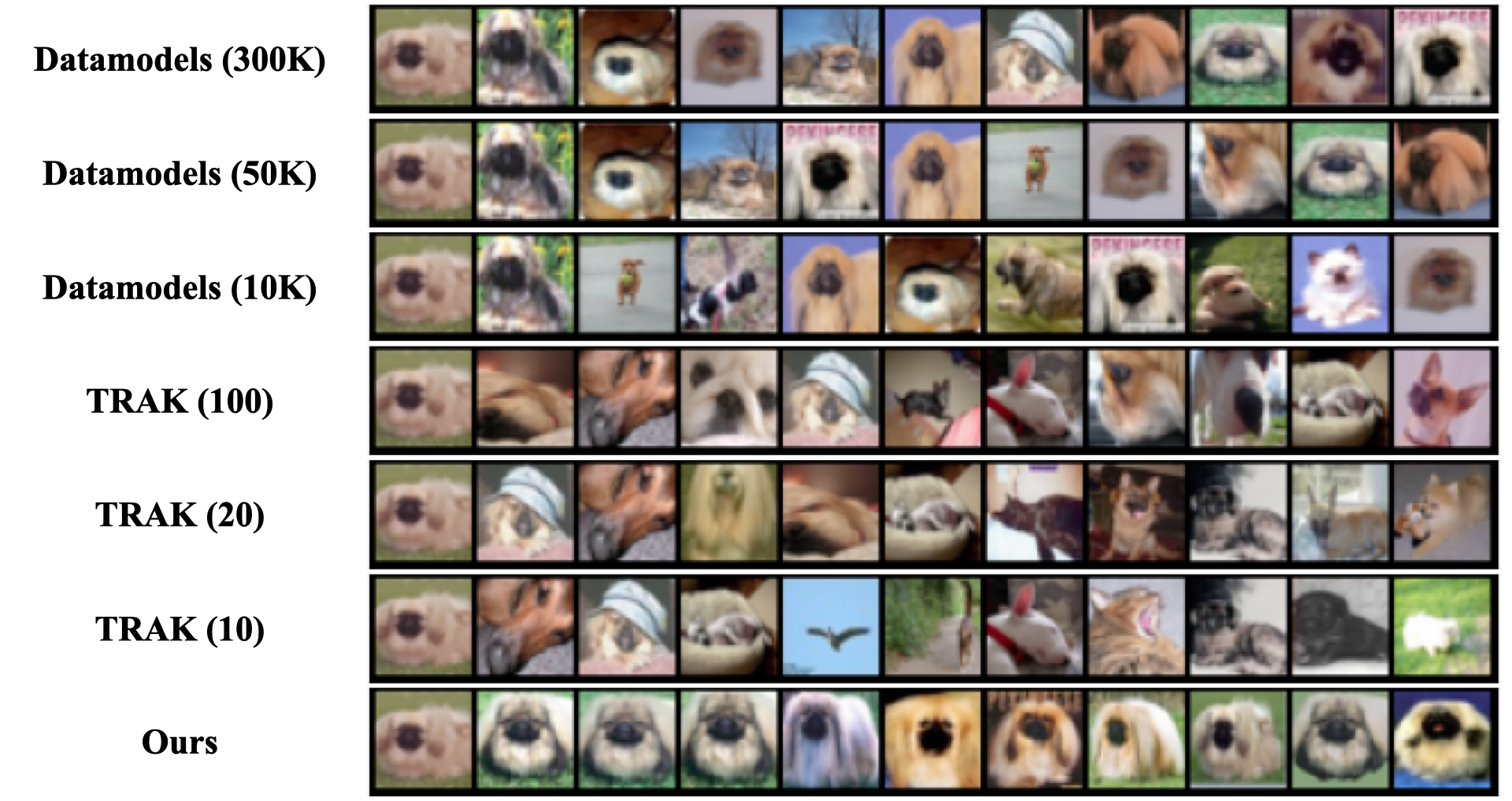}}
    }
    \caption{\textbf{Our attribution method consistently selects the most visually similar training images by design}. In each row, we plot the same target test image (Index 31), followed by ten most similar training images according to each attribution method. }
    \label{fig:most-similar-train-images-all-methods}
\end{figure}

In \cref{fig:most-similar-train-images-all-methods}, we plot the most similar training images according to Datamodels, TRAK, and our method. Given that our approach relies on comparing MoCo features from the same class as the target image, it makes sense that the closest training images are visually similar. On the other hand, the most similar training images found by Datamodels \citep{ilyas22datamodels} and TRAK \citep{park2023trak} show more variability. Despite the variability of most similar train images, Datamodels (300K) outperforms all other methods in the counterfactual tasks assessed in \cref{fig:k-leave-out}, hinting at the importance of additional contributing factors. Still, our method underscores the significant impact of relying solely on visual similarity, essentially showing that a significant fraction of data attribution can be achieved without knowledge of the learning algorithm, based only on knowledge of the training set.

% \tom{This example is acceptable, but TBH I think the visual similarity is more striking when you use a truck or plane or car example.  The image you chose here is kind of a mushy mess.}
\subsection{Other Related Work}
\label{sec:related}

% Discuss Datamodels, and TRAK in detail. Influence functions + related methods (Gradient Aggregated Similarity, TracIN) should be discussed as well. 
% Mention how these methods are useful for leave-out estimation. 

% How do changes to the training set influence model output? 
Data attribution methods should produce accurate counterfactual predictions about model outputs. Although a counterfactual can be addressed by retraining the model, employing this straightforward approach becomes impractical when dealing with large models and extensive datasets. To address this problem, data attribution methods perform various approximations.

% Tracing model output back to training data is related to the concept of \textit{empirical influence} \cite{hampel2011robust}, where one seeks to measure the dependence of an estimator on any one point in the sample.  

The seminal work on data attribution of \citet{koh2017understanding} proposes attribution via approximate \textit{influence functions}. More specifically, \citet{koh2017understanding} identify training samples most responsible for a given prediction by estimating the effect of removing or slightly modifying a single training sample. But being a first-order approximation, influence function estimates can vary wildly with changes to network architecture and training regularization \citep{basu2021influence}. Nevertheless, approximating influence functions is reasonably inexpensive and has recently also been attempted for multi-billion parameter models \cite{grosse_studying_2023}. 

Measuring empirical influence has also been attempted through construction of subsets of training data that include/exclude the target sample \citep{feldman2020neural}. In a related approach, TracIn \citep{pruthi2020estimating} and Gradient Aggregated Similarity (GAS) \citep{hammoudeh2022identifying, hammoudeh2022training} estimate the influence of each sample in training set $S$ on the test example $z_t$ by measuring the change in loss on $z_t$ from gradient updates of mini-batches. While TracIn can predict class margins reasonably well, the method struggles at estimating data support. Other methods for influence approximation include metrics based on representation similarity \citep{yeh2018representer, charpiat2019input}. Another related line of work has utilized Shapley values to ascribe value to data, but since Shapley values often require exponential time to compute, approximations have been proposed \citep{ghorbani2019data, jia2019towards}. In general, there seems to be a recurring tradeoff: methods that are computationally efficient tend to be less reliable, whereas sampling-based approaches are more effective but require training thousands (or even tens of thousands) of models.

% Maybe add a mention that our approach serves as a good prior for data attirbution

\section{Conclusion}
\label{sec:conclusion}
Data attribution approaches are computationally expensive and can be prone to inaccuracy. While these approaches exhibit promise and capability, their scalability to large-scale models remains uncertain.
Our work highlights the importance of visual similarity as a baseline for counterfactual estimation, providing valuable insights into data attribution. Our approach demonstrates scalability and accuracy, particularly in attributions for ImageNet, where it outperforms other state-of-the-art methods while maintaining manageable compute and storage requirements. Remarkably, our approach achieves these results without any reliance on training setup details, target model parameters, or architectural specifics. Our work shows that strong data attribution can be achieved solely based on knowledge of the training set.

% Gigantic TODO :(
% \input{figures/compare_supports}

\section{Acknowledgments and Disclosure of Funding}
\label{sec:acknowledgments}
Vasu Singla was supported in part
by the National Science Foundation under grant number IIS-2213335. Pedro Sandoval-Segura is supported by a National Defense Science and Engineering Graduate (NDSEG) Fellowship. Tom Goldstein was supported by the ONR MURI program, the AFOSR MURI program, the National Science Foundation (IIS-2212182), and by the NSF TRAILS Institute (2229885).

\bibliography{main}
\bibliographystyle{iclr2024_conference}
% \section*{References}

% \medskip

% {
% \small
% \bibliographystyle{plain}
% \bibliography{references}
% }

%%%%%%%%%%%%%%%%%%%%%%%%%%%%%%%%%%%%%%%%%%%%%%%%%%%%%%%%%%%%

\newpage

\appendix

\section{Appendix}

\subsection{Compute Time and Storage Requirements }
\label{app:compute-description}

For our compute time estimates, we use NVIDIA RTX A6000 GPUs and 4 CPU cores. We describe how we estimate the wall-clock time, and storage requirements for each method below -

\begin{itemize}
    \item \textbf{Datamodels:} We only take into account the storage and compute cost of training models. The additional cost of estimating datamodels from the trained models, requires solving linear regression whose computational costs are negligible compared to training the models. For compute and storage requirement estimates, we train $100$ ResNet-9 models on random $50\%$ subsets of CIFAR-10 and extrapolate to estimate the training time and storage required for 10,000 and 50,000 models shown in \cref{fig:compute-efficiency}. 
    \item \textbf{TRAK:} We use the authors' original code \footnote{https://github.com/MadryLab/trak} to train, and compute the projected gradients for CIFAR-10 using ResNet-9 Models using a projection dimension of 20480. For storage requirements, we take into account storage used by model weights, and the projected gradients. The results in \cref{fig:compute-efficiency}, show the compute and storage using 10, 20 and 100 models.
    \item \textbf{Ours:} We use Lightly library \footnote{https://github.com/lightly-ai/lightly} benchmark code to train a MoCo model using a ResNet-18 backbone on CIFAR-10 for 800 epochs. The results in \cref{fig:compute-efficiency} show the wall-clock training time for the model, and extracting the features from CIFAR-10 and the storage requirements for model weights.
\end{itemize}

To calculate the storage requirements, we factor in the storage space necessary for retaining the trained model weights, as they are essential for computing influence on new validation samples across all attribution methods.

\subsection{Additional Self-Supervised Features}
\label{app:additional-self-supervised}

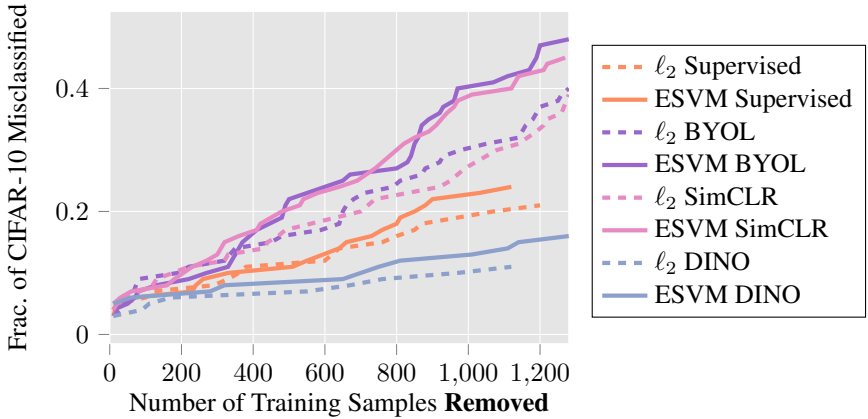
\begin{figure}[ht]
    \centering
    \begin{tikzpicture}
    \centering
    \definecolor{darkgray141160203}{RGB}{141,160,203}
    \definecolor{dimgray85}{RGB}{85,85,85}
    \definecolor{gainsboro229}{RGB}{229,229,229}
    \definecolor{lightgray204}{RGB}{204,204,204}
    \definecolor{mediumaquamarine102194165}{RGB}{102,194,165}
    \definecolor{orchid231138195}{RGB}{231,138,195}
    \definecolor{salmon25214198}{RGB}{252,141,98}
    \definecolor{amethyst}{rgb}{0.6, 0.4, 0.8}
    \definecolor{bleudefrance}{rgb}{0.19, 0.55, 0.91}
    \definecolor{blush}{rgb}{0.87, 0.36, 0.51}
    \definecolor{brilliantrose}{rgb}{1.0, 0.33, 0.64}
    \begin{groupplot}[
        group style={group size= 1 by 1, horizontal sep=0cm, vertical sep=0cm},
        height={6cm},
        width=.55\linewidth]
    \nextgroupplot[
        xlabel={Number of Training Samples \textbf{Removed}},
        ylabel={Frac. of CIFAR-10 Misclassified},
        legend pos={north east},
        axis background/.style={fill=gainsboro229},
        axis line style={white},
        legend cell align={left},
        legend columns=1,
        legend style={
        fill opacity=0.8,
        draw opacity=1,
        text opacity=1,
        at={(1.05,0.9)},
        anchor=north west,
        draw=black,
        fill=none,
        },
        ymajorgrids=true,
        mark size=1.0pt,
        xmin=0, xmax=1280,
        xmajorgrids, xminorgrids,
        x grid style={white},
        y grid style={white},
        tick align=outside,
        tick pos=left,
    ]

    % \addplot[ultra thick, color=mediumaquamarine102194165] table [x=x, y=y, col sep=comma] {data/brittleness_remove/dmodel.csv};
    % \addlegendentry{Datamodels ($300$k models)}
    
    % L2 ResNet-9
    \addplot[ultra thick, color=salmon25214198, dashed] table [x=x, y=y, col sep=comma] {data/brittleness_remove/L2-ResNet9.csv};
    \addlegendentry{$\ell_2$ Supervised}

    % ESVM ResNet-9
    \addplot[ultra thick, color=salmon25214198] table [x=x, y=y, col sep=comma] {data/brittleness_remove/ESVM-ResNet9.csv};
    \addlegendentry{ESVM Supervised}

    \addplot[ultra thick, color=amethyst, dashed] table [x=x, y=y, col sep=comma] {data/brittleness_remove/L2-BYOL.csv};
    \addlegendentry{$\ell_2$ BYOL}

    \addplot[ultra thick, color=amethyst] table [x=x, y=y, col sep=comma] {data/brittleness_remove/ESVM-BYOL.csv};
    \addlegendentry{ESVM BYOL}

    \addplot[ultra thick, color=orchid231138195, dashed] table [x=x, y=y, col sep=comma] {data/brittleness_remove/L2-SimCLR.csv};
    \addlegendentry{$\ell_2$ SimCLR}

    \addplot[ultra thick, color=orchid231138195] table [x=x, y=y, col sep=comma] {data/brittleness_remove/ESVM-SimCLR.csv};
    \addlegendentry{ESVM SimCLR}

    \addplot[ultra thick, color=darkgray141160203, dashed] table [x=x, y=y, col sep=comma] {data/brittleness_remove/L2-DINO.csv};
    \addlegendentry{$\ell_2$ DINO}

    \addplot[ultra thick, color=darkgray141160203] table [x=x, y=y, col sep=comma] {data/brittleness_remove/ESVM-DINO.csv};
    \addlegendentry{ESVM DINO}
    
    \end{groupplot}
\end{tikzpicture}
    \caption{We estimate data removal support for 100 random CIFAR-10 test samples and plot the CDF of estimates.}
\label{fig:k-leave-out-l2-vs-esvm-appendix}
\end{figure}

\begin{figure}[ht]
    \centering
    \begin{tikzpicture}
    \centering
    \definecolor{darkgray141160203}{RGB}{141,160,203}
    \definecolor{dimgray85}{RGB}{85,85,85}
    \definecolor{gainsboro229}{RGB}{229,229,229}
    \definecolor{lightgray204}{RGB}{204,204,204}
    \definecolor{mediumaquamarine102194165}{RGB}{102,194,165}
    \definecolor{orchid231138195}{RGB}{231,138,195}
    \definecolor{salmon25214198}{RGB}{252,141,98}
    \definecolor{amethyst}{rgb}{0.6, 0.4, 0.8}
    \definecolor{bleudefrance}{rgb}{0.19, 0.55, 0.91}
    \definecolor{blush}{rgb}{0.87, 0.36, 0.51}
    \definecolor{brilliantrose}{rgb}{1.0, 0.33, 0.64}
    \begin{groupplot}[
        group style={group size= 1 by 1, horizontal sep=0cm, vertical sep=0cm},
        height={6cm},
        width=.55\linewidth]
    \nextgroupplot[
        xlabel={Number of Training Samples \textbf{Mislabeled}},
        ylabel={Frac. of CIFAR-10 Misclassified},
        legend pos={north east},
        axis background/.style={fill=gainsboro229},
        axis line style={white},
        legend cell align={left},
        legend columns=1,
        legend style={
        fill opacity=0.8,
        draw opacity=1,
        text opacity=1,
        at={(1.05,0.9)},
        anchor=north west,
        draw=black,
        fill=none,
        },
        ymajorgrids=true,
        mark size=1.5pt,
        % no marks,
        xmin=10, xmax=1280,
        xmajorgrids, xminorgrids,
        x grid style={white},
        y grid style={white},
        tick align=outside,
        tick pos=left,
    ]

    % \addplot[semithick, color=mediumaquamarine102194165, mark=*] table [x=x, y=y, col sep=comma] {data/brittleness_flip/dmodel.csv};
    % \addlegendentry{Datamodels ($300$k models)}

    \addplot[ultra thick, color=salmon25214198, dashed] table [x=x, y=y, col sep=comma] {data/brittleness_flip/L2-ResNet9.csv};
    \addlegendentry{$\ell_2$ Supervised}

    \addplot[ultra thick, color=salmon25214198] table [x=x, y=y, col sep=comma] {data/brittleness_flip/ESVM-ResNet9.csv};
    \addlegendentry{ESVM Supervised}

    \addplot[ultra thick, color=bleudefrance, dashed] table [x=x, y=y, col sep=comma] {data/brittleness_flip/L2-MoCo.csv};
    \addlegendentry{$\ell_2$ MoCo}

    \addplot[ultra thick, color=bleudefrance] table [x=x, y=y, col sep=comma] {data/brittleness_flip/ESVM-MoCo.csv};
    \addlegendentry{ESVM MoCo}

    \addplot[ultra thick, color=amethyst, dashed] table [x=x, y=y, col sep=comma] {data/brittleness_flip/L2-BYOL.csv};
    \addlegendentry{$\ell_2$ BYOL}

    \addplot[ultra thick, color=amethyst] table [x=x, y=y, col sep=comma] {data/brittleness_flip/ESVM-BYOL.csv};
    \addlegendentry{ESVM BYOL}

    \addplot[ultra thick, color=orchid231138195, dashed] table [x=x, y=y, col sep=comma] {data/brittleness_flip/L2-SimCLR.csv};
    \addlegendentry{$\ell_2$ SimCLR}

    \addplot[ultra thick, color=orchid231138195] table [x=x, y=y, col sep=comma] {data/brittleness_flip/ESVM-SimCLR.csv};
    \addlegendentry{ESVM SimCLR}

    \addplot[ultra thick, color=darkgray141160203, dashed] table [x=x, y=y, col sep=comma] {data/brittleness_flip/L2-DINO.csv};
    \addlegendentry{$\ell_2$ DINO}

    \addplot[ultra thick, color=darkgray141160203] table [x=x, y=y, col sep=comma] {data/brittleness_flip/ESVM-DINO.csv};
    \addlegendentry{ESVM DINO}

    \end{groupplot}
\end{tikzpicture}
    \caption{We estimate data mislabel support for 100 random CIFAR-10 test samples and plot the CDF of estimates.}
\label{fig:k-flip-l2-vs-esvm-appendix}
\end{figure}
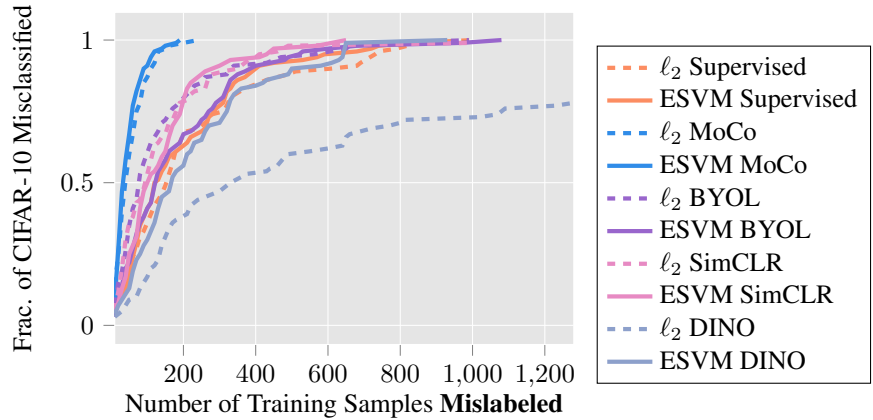

In addition to utilizing features from MoCo in \cref{sec:our-approach-and-baselines}, we test our choice of distance function on ResNet-18 features from other self-supervised learning (SSL) methods trained on CIFAR-10. In particular, we evaluate BYOL \citep{grill2020bootstrap}, SimCLR \citep{chen2020simple}, and DINO \citep{caron2021emerging} at estimating data removal support in \cref{fig:k-leave-out-l2-vs-esvm-appendix} and mislabel support in \cref{fig:k-flip-l2-vs-esvm-appendix}. With the exception of DINO, self-supervised features from BYOL and SimCLR outperform the supervised baseline at estimating data removal support. Additionally, we see that in all cases using ESVM distance is more effective than using $\ell_2$ distance to compare features. 
% We suspect the DINO framework didn't perform well for our approach due to the smaller dataset, 

\subsubsection{Self-Supervised ImageNet Features}

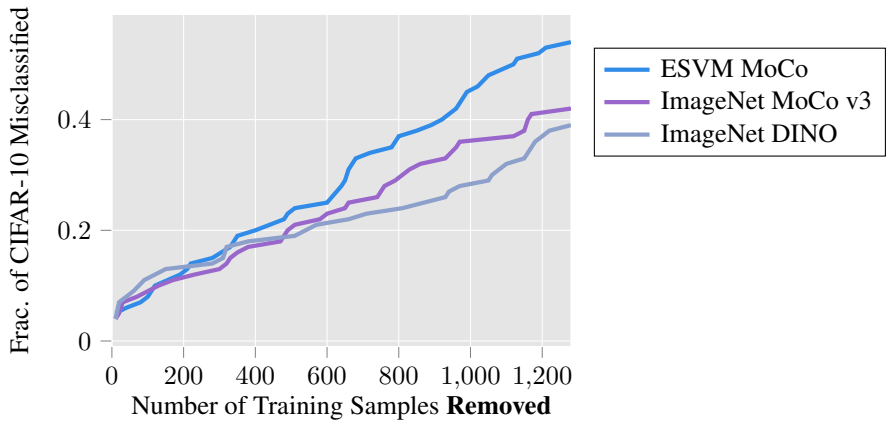
\begin{figure}[!htb]
    \centering
    \begin{tikzpicture}
    \centering
    \definecolor{darkgray141160203}{RGB}{141,160,203}
    \definecolor{dimgray85}{RGB}{85,85,85}
    \definecolor{gainsboro229}{RGB}{229,229,229}
    \definecolor{lightgray204}{RGB}{204,204,204}
    \definecolor{mediumaquamarine102194165}{RGB}{102,194,165}
    \definecolor{orchid231138195}{RGB}{231,138,195}
    \definecolor{salmon25214198}{RGB}{252,141,98}
    \definecolor{amethyst}{rgb}{0.6, 0.4, 0.8}
    \definecolor{bleudefrance}{rgb}{0.19, 0.55, 0.91}
    \definecolor{blush}{rgb}{0.87, 0.36, 0.51}
    \definecolor{brilliantrose}{rgb}{1.0, 0.33, 0.64}
    \begin{groupplot}[
        group style={group size= 1 by 1, horizontal sep=0cm, vertical sep=0cm},
        height={6cm},
        width=.55\linewidth]
    \nextgroupplot[
        xlabel={Number of Training Samples \textbf{Removed}},
        ylabel={Frac. of CIFAR-10 Misclassified},
        legend pos={north east},
        axis background/.style={fill=gainsboro229},
        axis line style={white},
        legend cell align={left},
        legend columns=1,
        legend style={
        column sep=0.05cm,
        fill opacity=0.8,
        draw opacity=1,
        text opacity=1,
        at={(1.05, 0.9)},
        anchor=north west,
        fill=none,
        draw=black,
        },
        ymajorgrids=true,
        mark size=1.0pt,
        xmin=0, xmax=1280,
        xmajorgrids, xminorgrids,
        x grid style={white},
        y grid style={white},
        tick align=outside,
        tick pos=left,
    ]

    % Datamodels
    % \addplot[ultra thick, color=mediumaquamarine102194165] table [x=x, y=y, col sep=comma] {data/brittleness_remove/dmodel.csv};
    % \addlegendentry{Datamodels ($300$k models)}
    
    % ESVM MoCo ResNet-9
    \addplot[ultra thick, color=bleudefrance] table [x=x, y=y, col sep=comma] {data/brittleness_remove/ESVM-MoCo-ResNet9.csv};
    \addlegendentry{ESVM MoCo}

    % ImageNet MoCo v3 (ViT/B16 300epoch from vissl library)
    \addplot[ultra thick, color=amethyst] table [x=x, y=y, col sep=comma] {data/brittleness_remove/ESVM-ImageNetMoCov3-ViT-B16-300epoch.csv};
    \addlegendentry{ImageNet MoCo v3}   

    % ImageNet DINO (ViT/S16 300epoch from vissl library)
    \addplot[ultra thick, color=darkgray141160203] table [x=x, y=y, col sep=comma] {data/brittleness_remove/ESVM-ImageNetDINO-ViT-S16-300epoch.csv};
    \addlegendentry{ImageNet DINO}    
    
    \end{groupplot}
\end{tikzpicture}
    \caption{ImageNet features from MoCo v3 and DINO are able to perform very well despite using different architecture (\ie ViT), dataset (\ie ImageNet), and learning objectives (\ie SSL) from the system that we are trying to attribute predictions for: a ResNet-9 trained normally on CIFAR-10.}
\label{fig:k-leave-out-imagenet-ssl-features}
\end{figure}

We also consider using ImageNet features from MoCo v3 \citep{chen2021mocov3} and DINO \citep{caron2021emerging} to estimate data removal support in \cref{fig:k-leave-out-imagenet-ssl-features}. We use publicly available MoCo v3 and DINO checkpoints from the \texttt{vissl} library's model zoo \citep{goyal2021vissl}. It is worth noting that this approach places significant emphasis our primary hypothesis, which asserts the importance of visual similarity in data attribution. Utilizing ImageNet features means that the dataset, architecture, and learning objectives are \textit{completely different} from the system we are trying to attribute predictions for: a ResNet-9 trained normally on CIFAR-10. This is in contrast to our main method (ESVM MoCo) which utilizes a ResNet-18 architecture and the CIFAR-10 dataset. 

\subsection{Additional Justification for Chosen Subset of Train Images}
\label{app:chosen-subset-of-train}

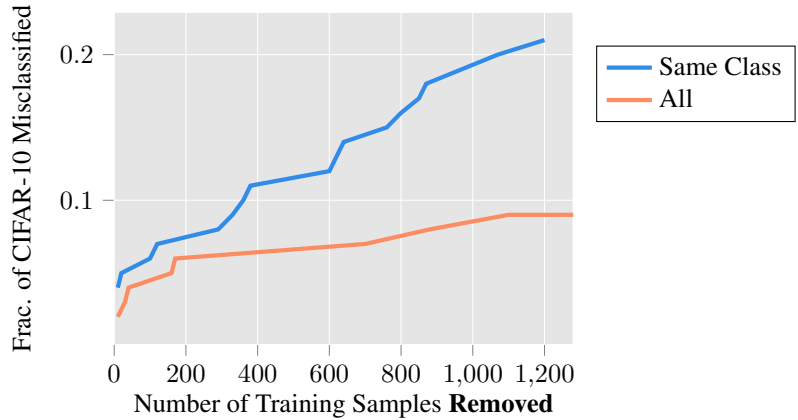
\begin{figure}[ht]
    \centering
    \begin{tikzpicture}
    \centering
    \definecolor{darkgray141160203}{RGB}{141,160,203}
    \definecolor{dimgray85}{RGB}{85,85,85}
    \definecolor{gainsboro229}{RGB}{229,229,229}
    \definecolor{lightgray204}{RGB}{204,204,204}
    \definecolor{mediumaquamarine102194165}{RGB}{102,194,165}
    \definecolor{orchid231138195}{RGB}{231,138,195}
    \definecolor{salmon25214198}{RGB}{252,141,98}
    \definecolor{amethyst}{rgb}{0.6, 0.4, 0.8}
    \definecolor{bleudefrance}{rgb}{0.19, 0.55, 0.91}
    \definecolor{blush}{rgb}{0.87, 0.36, 0.51}
    \definecolor{brilliantrose}{rgb}{1.0, 0.33, 0.64}
    \begin{groupplot}[
        group style={group size= 1 by 1, horizontal sep=0cm, vertical sep=0cm},
        height={6cm},
        width=.55\linewidth]
    \nextgroupplot[
        xlabel={Number of Training Samples \textbf{Removed}},
        ylabel={Frac. of CIFAR-10 Misclassified},
        legend pos={north east},
        axis background/.style={fill=gainsboro229},
        axis line style={white},
        legend cell align={left},
        legend columns=1,
        legend style={
        fill opacity=0.8,
        draw opacity=1,
        text opacity=1,
        at={(1.05,0.9)},
        anchor=north west,
        draw=black,
        fill=none,
        },
        ymajorgrids=true,
        mark size=1.0pt,
        xmin=0, xmax=1280,
        xmajorgrids, xminorgrids,
        x grid style={white},
        y grid style={white},
        tick align=outside,
        tick pos=left,
    ]

    \addplot[ultra thick, color=bleudefrance] table [x=x, y=y, col sep=comma] {data/brittleness_remove/L2-ResNet9.csv};
    \addlegendentry{Same Class}

    \addplot[ultra thick, color=salmon25214198] table [x=x, y=y, col sep=comma] {data/brittleness_remove/L2-ResNet9-all.csv};
    \addlegendentry{All}

    \end{groupplot}
\end{tikzpicture}
    \caption{Choosing removal support from all training images is less effective than selecting from the same class as the target image.}
\label{fig:k-flip-l2-all-vs-same-class-appendix}
\end{figure}

For a target sample $z_t$, data attribution approaches rank the training samples based on decreasing order of positive influence on $z_t$. For our method, a design choice was whether to rank training samples from all classes or from a selected subset of the training data. One reasonable subset was to select training samples from the same class as the target test sample. In \cref{fig:k-flip-l2-all-vs-same-class-appendix}, we show that selecting from the same class is more effective when estimating britteness scores. We maintain this choice for all our experiments.

\subsection{Additional Justification for Distance Function}

In \cref{sec:design-choices}, we describe choices for measuring similarity of embeddings: Euclidean distance, cosine distance, and our selection of Exemplar SVM. However, there are a range of other metrics that have been evaluated in prior work. By no means have we exhausted the space of possible metrics, but it relevant to look at recommendations by related work.

\subsubsection{Gradient Cosine Similarity}
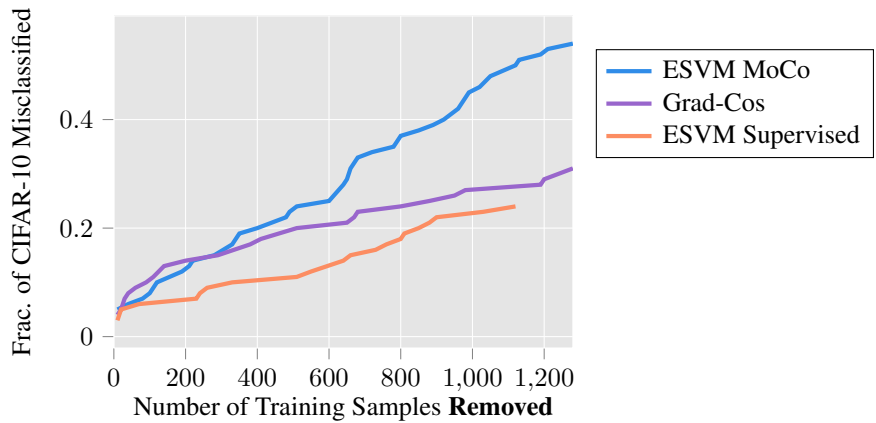
\begin{figure}[!htb]
    \centering
    \begin{tikzpicture}
    \centering
    \definecolor{darkgray141160203}{RGB}{141,160,203}
    \definecolor{dimgray85}{RGB}{85,85,85}
    \definecolor{gainsboro229}{RGB}{229,229,229}
    \definecolor{lightgray204}{RGB}{204,204,204}
    \definecolor{mediumaquamarine102194165}{RGB}{102,194,165}
    \definecolor{orchid231138195}{RGB}{231,138,195}
    \definecolor{salmon25214198}{RGB}{252,141,98}
    \definecolor{amethyst}{rgb}{0.6, 0.4, 0.8}
    \definecolor{bleudefrance}{rgb}{0.19, 0.55, 0.91}
    \definecolor{blush}{rgb}{0.87, 0.36, 0.51}
    \definecolor{brilliantrose}{rgb}{1.0, 0.33, 0.64}
    \begin{groupplot}[
        group style={group size= 1 by 1, horizontal sep=0cm, vertical sep=0cm},
        height={6cm},
        width=.55\linewidth]
    \nextgroupplot[
        xlabel={Number of Training Samples \textbf{Removed}},
        ylabel={Frac. of CIFAR-10 Misclassified},
        legend pos={north east},
        axis background/.style={fill=gainsboro229},
        axis line style={white},
        legend cell align={left},
        legend columns=1,
        legend style={
        column sep=0.05cm,
        fill opacity=0.8,
        draw opacity=1,
        text opacity=1,
        at={(1.05, 0.9)},
        anchor=north west,
        fill=none,
        draw=black,
        },
        ymajorgrids=true,
        mark size=1.0pt,
        xmin=0, xmax=1280,
        xmajorgrids, xminorgrids,
        x grid style={white},
        y grid style={white},
        tick align=outside,
        tick pos=left,
    ]

    % Datamodels
    % \addplot[ultra thick, color=mediumaquamarine102194165] table [x=x, y=y, col sep=comma] {data/brittleness_remove/dmodel.csv};
    % \addlegendentry{Datamodels ($300$k models)}
    
    % ESVM MoCo ResNet-9
    \addplot[ultra thick, color=bleudefrance] table [x=x, y=y, col sep=comma] {data/brittleness_remove/ESVM-MoCo-ResNet9.csv};
    \addlegendentry{ESVM MoCo}

    % Grad-Cos ResNet-9
    \addplot[ultra thick, color=amethyst] table [x=x, y=y, col sep=comma] {data/brittleness_remove/Grad-Cos-ResNet9-NoFilter.csv};
    \addlegendentry{Grad-Cos}    

    % ESVM ResNet-9
    \addplot[ultra thick, color=salmon25214198] table [x=x, y=y, col sep=comma] {data/brittleness_remove/ESVM-ResNet9.csv};
    \addlegendentry{ESVM Supervised}

    % \addplot[ultra thick, color=black]{}
    \end{groupplot}
\end{tikzpicture}
    \caption{While comparing images with Gradient Cosine Similarity (using a supervised ResNet-9) is better than ESVM on supervised features, it still lags behind our main method (ESVM MoCo).}
\label{fig:k-leave-out-grad-cos}
\end{figure}

\cite{hanawa2021evaluation} define a set of tests that a similarity metric should satisfy and find that gradient cosine similarity (Grad-Cos) is the only one that passes all tests. Given that Grad-Cos is their overall recommendation for measuring similarity, we evaluate data removal support on CIFAR-10 in \cref{fig:k-leave-out-grad-cos}. Note that unlike other methods considered, we do not filter images to be of the same class as the target because Grad-Cos already provides a higher ranking to images from the same target class. While we find that Grad-Cos is better than ESVM comparison of supervised features, it still lags behind our main method (ESVM MoCo) from \cref{sec:our-approach-and-baselines}. Interestingly, in the low data support regime, where fewer than 200 training samples can be removed to misclassify, Grad-Cos is more effective than ESVM MoCo. 

\subsubsection{Human Visual Similarity \& DreamSim}

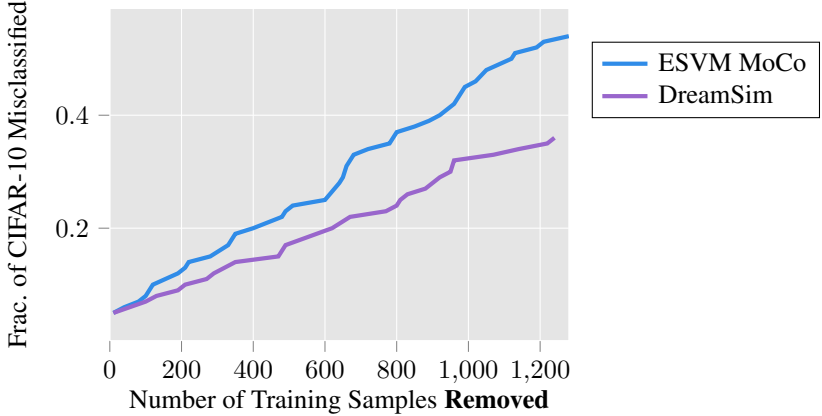
\begin{figure}[!htb]
    \centering
    \begin{tikzpicture}
    \centering
    \definecolor{darkgray141160203}{RGB}{141,160,203}
    \definecolor{dimgray85}{RGB}{85,85,85}
    \definecolor{gainsboro229}{RGB}{229,229,229}
    \definecolor{lightgray204}{RGB}{204,204,204}
    \definecolor{mediumaquamarine102194165}{RGB}{102,194,165}
    \definecolor{orchid231138195}{RGB}{231,138,195}
    \definecolor{salmon25214198}{RGB}{252,141,98}
    \definecolor{amethyst}{rgb}{0.6, 0.4, 0.8}
    \definecolor{bleudefrance}{rgb}{0.19, 0.55, 0.91}
    \definecolor{blush}{rgb}{0.87, 0.36, 0.51}
    \definecolor{brilliantrose}{rgb}{1.0, 0.33, 0.64}
    \begin{groupplot}[
        group style={group size= 1 by 1, horizontal sep=0cm, vertical sep=0cm},
        height={6cm},
        width=.55\linewidth]
    \nextgroupplot[
        xlabel={Number of Training Samples \textbf{Removed}},
        ylabel={Frac. of CIFAR-10 Misclassified},
        legend pos={north east},
        axis background/.style={fill=gainsboro229},
        axis line style={white},
        legend cell align={left},
        legend columns=1,
        legend style={
        column sep=0.05cm,
        fill opacity=0.8,
        draw opacity=1,
        text opacity=1,
        at={(1.05, 0.9)},
        anchor=north west,
        fill=none,
        draw=black,
        },
        ymajorgrids=true,
        mark size=1.0pt,
        xmin=0, xmax=1280,
        xmajorgrids, xminorgrids,
        x grid style={white},
        y grid style={white},
        tick align=outside,
        tick pos=left,
    ]

    % Datamodels
    % \addplot[ultra thick, color=mediumaquamarine102194165] table [x=x, y=y, col sep=comma] {data/brittleness_remove/dmodel.csv};
    % \addlegendentry{Datamodels ($300$k models)}
    
    % ESVM MoCo ResNet-9
    \addplot[ultra thick, color=bleudefrance] table [x=x, y=y, col sep=comma] {data/brittleness_remove/ESVM-MoCo-ResNet9.csv};
    \addlegendentry{ESVM MoCo}

    % Grad-Cos ResNet-9
    \addplot[ultra thick, color=amethyst] table [x=x, y=y, col sep=comma] {data/brittleness_remove/DreamSim-OpenCLIP.csv};
    \addlegendentry{DreamSim}    

    % \addplot[ultra thick, color=black]{}
    \end{groupplot}
\end{tikzpicture}
    \caption{We use DreamSim (OpenCLIP-ViTB/32) to select data removal support on CIFAR-10.}
\label{fig:k-leave-out-dreamsim-openclip}
\end{figure}

\cite{fu2023dreamsim} study perceptual metrics and find that large vision models like OpenCLIP \citep{cherti2023reproducible} and DINO \citep{caron2021emerging} are more aligned with human perceptual judgements than other learned metrics like LPIPS \citep{zhang2018unreasonable} and DISTS \citep{ding2020image}. They further improve performance of OpenCLIP and DINO by finetuning with LoRA \citep{hu2021lora} on a dataset of human two-alternative forced choice (2AFC) judgments, called NIGHTS. The best approach on the dataset uses an ensemble of DINO, CLIP, and OpenCLIP features and is called DreamSim. While the ensemble gets $96.2\%$ accuracy on NIGHTS, only utilizing OpenCLIP (with LoRA) gets $95.5\%$ and is $3\times$ faster. Hence, we use this metric in our data removal support evaluation. Here, we also select images from the same training class. In \cref{fig:k-leave-out-dreamsim-openclip}, for every target image, we select the closest training images according to DreamSim to remove. Surprisingly, DreamSim does not improve over our approach using ESVM MoCo.

\subsection{Computing Data Support}
\label{app:computing-support}

% 1. \vasu{Add bisection search algorithm, and discuss why it works, linearity assumption}

We use bisection search to estimate data support. The use of bisection search is supported by the observation that several data attribution approaches are additive \citep{park2023trak}, where the importance of a subset of training samples is defined as the sum of each of the samples in the subset.  To compute data removal support, we remove $M$ samples (chosen using each attribution method) from the training data and log whether the resulting model misclassifies the target sample. For data mislabeling support, we mislabel $M$ samples (chosen using each attribution method) from the training data and assign a new label corresponding to the highest incorrect logit.

A detailed summary of our bisection search is in \cref{algo:bisection-search-data-support}. A key step is $\mathrm{CounterfactualTest}(f, S, I_{\rm attr}[:M])$ which returns the average classification of $N_{\rm test}$ independent training runs where $f_{\theta}$ is trained on the subset $R = \{z_i | z_i \in S \text{ and } i \notin I_{\rm attr}[:M]\}$. In other words, for computing data removal support, $f_\theta$ is trained on a subset of $S$ that does not include the first $M$ indices of $I_{\rm attr}$. For computing mislabeling data support, the only difference is that rather than removing the first $M$ indices of $I_{\rm attr}$, we relabel those samples with the class of the highest incorrect-class logit, following \citep{ilyas22datamodels}.

% For computing mislabeling data support, $\mathrm{CounterfactualTest}(f, S, I_{\rm attr}[:M])$ returns the average classification of $N_{\rm test}$ independent training runs where $f_{\theta}$ is trained on the subset $R = \{z_i | z_i \in S \text{ and } i \notin I_{\rm attr}[:M]\} \cup \{ z_i = (x_i, \Tilde{y}_i) | i \in I_{\rm attr}[:M] \text{ and } \Tilde{y} \text{ is most-likely incorrect class}\}$, so $M$ samples are mislabeled.

\renewcommand{\algorithmicrequire}{\textbf{Input: }}
\renewcommand{\algorithmicensure}{\textbf{Output: }}

\begin{algorithm}[h]
\caption{Bisection Search for Computing Data Support}
\label{algo:bisection-search-data-support}
\algorithmicrequire Target sample, $z_t = (x_t, y_t)$ \\
\algorithmicrequire Training set, $S$, and a list of top $k$ training set indices $I_{\rm attr}$ ordered by the attribution method $\tau(z, S)$ \\
\algorithmicrequire Model $f_\theta$ \\
\algorithmicrequire Search budget, $N_{\mathrm{budget}}$ \\
\algorithmicrequire Number of times to test classification, $N_{\mathrm{test}}$ \\
\algorithmicensure $N_{\rm support}$, size of the smallest training subset $R \subset S$ such that $f_{\theta}$ misclassifies $x_t$ on average
\begin{algorithmic}[1]
    \State L $\gets 0$ 
    \State H $\gets |I_{\rm attr}|$ 
    \State M $\gets H$ 
    \State $C_{\rm avg} \gets \mathrm{CounterfactualTest}(f, S, I_{\rm attr}[:M])$
    \If{$C_{\rm avg} > 0.5$}
        \State \Return -1 \Comment{$N_{\rm support}$ is larger than $k$}
    \EndIf
    % $ Train $N_{\mathrm{test}}$ independent models $f_{\theta_i}$, and compute average correctness
    \State $N_{\rm support} \gets M$
    \While{$N_{\mathrm{budget}} > 0$}
        \State $N_{\mathrm{budget}} \gets N_{\mathrm{budget}} - 1$
        \State $M \gets (L + H) / 2$
        \State $C_{\rm avg} \gets \mathrm{CounterfactualTest}(f, S, I_{\rm attr}[:M])$
        \If{$C_{\rm avg} > 0.5$}
            \State $L \gets M$
        \Else 
            \State $H \gets M$
            \State $N_{\rm support} \gets \min(M, N_{\rm support}) $
        \EndIf
    \EndWhile
    \State \Return $N_{\rm support}$
\end{algorithmic}
\end{algorithm}

For bisection search across all attribution methods, we use a search budget of 7. For the CIFAR-10 data brittleness metrics, we aggregate predictions over 5 independently trained models. Thus, to evaluate a single validation sample, we train 35 models  (7 budget $\times$ 5 models) for a total of 3500 (35 $\times$ 100 samples) models for a data brittleness metric. On ImageNet, we don't aggregate predictions and only train a single model. Hence, to evaluate a single validation sample on Imagenet, we train 7 models per sample, and a total of 210 models for evaluating a data brittleness metric. Due to the large training cost on ImageNet, we only show results for data removal support. We explicitly point out that these costs are incurred only for analysis of these data attribution methods (see \cref{sec:methodology}). Our attribution approach is in comparison, extremely cheap to compute.

\subsection{Linear Datamodeling Score}
\label{app:lds-scores}

Let $\tau(z, S): \mathcal{Z} \times \mathcal{Z}^n \rightarrow \mathbb{R}^n$ be a data attribution method that, for any sample $z \in \mathcal{Z}$ and a training set $S$ assigns a score to every training sample indicating its importance to the model output. Consider a training set $S = \{z_1, z_2 \dots z_n\}$, and a model output function $f_{\theta}(z)$. Let $\{S_1, ..., S_m | S_i \subset S \}$ be $m$ random subsets of the training set $S$, each of size $\alpha \cdot n$ for some $\alpha \in (0,1)$. The linear datamodeling score (LDS) is defined as:

\begin{equation}
    \mathrm{LDS}(\tau(z, S)) = \rho(\{ f_{\theta(S_j)}(z) \enspace|\enspace j \in [m]\}, \{ \tau(z, S)\cdot \1_{S_j} \enspace|\enspace  j \in [m]\})
\end{equation}

\noindent
where $\rho$ denotes Spearman rank correlation \citep{kokoska2000crc},  $\theta(S_j)$ denotes model parameters after training on subset $S_j$, and $\1_{S_j}$ is the indicator vector of the subset $S_j$ . Unlike data brittleness metrics, LDS accounts for samples with positive as well as negative influence. 

To compute LDS scores, for our model output function $f_{\theta}(z)$, we use the correct class margin. This is defined as: 
$$
    f_{\theta}(z) = (\text{logit for correct class}) - (\text{highest incorrect logit})
$$

\begin{table}[ht]
    \centering
    \begin{tabular}{ccc}
\toprule
\multicolumn{1}{l}{} & Models Used & LDS Scores \\
\midrule
\multirow{3}{*}{Datamodels} & 300,000 & 0.56 \\ 
 & 50,000 & 0.43 \\
 & 10,000 & 0.24 \\
 \midrule
\multirow{4}{*}{TRAK} & 100 & 0.22 \\
 & 20 & 0.15 \\
 & 10 & 0.12 \\
 & 5 & 0.08 \\
 \midrule
Ours & 1 & 0.08 \\
\bottomrule
\end{tabular}
% \label{tab:lds-scores}
% \captionof{table}{We compare LDS scores for our approach with other baselines on CIFAR-10. Our proposed approach can perform equivalent to TRAK with $5$ models.}
    \vspace{2mm}
    \caption{We compare LDS scores for our approach with other baselines on CIFAR-10. Our proposed approach can perform equivalent to TRAK with $5$ models.}
    \label{tab:lds-scores}
\end{table}

Our approach cannot directly be applied to compute LDS scores, as for a validation sample $z_t$ we only focus on training samples with the most positive impact. We propose a simple modification to our approach. We assign a score to each training data based on the inverse of signed $l_2$ distance. The sign is based on whether the label for the training sample matches $z_t$. We then threshold our scores, such that all scores beyond the top-$5\%$ are zero leading to sparser attribution scores. The sparsity prior has been shown to be effective for data attribution \citep{ilyas22datamodels, park2023trak}.

In \cref{tab:lds-scores}, we present a comparison of LDS scores using our baseline approach, TRAK and Datamodels.  Although our baseline was not initially designed for direct LDS score approximation, a simple adaptation demonstrates comparable performance to TRAK (5) on CIFAR-10. TRAK with a larger ensemble of models can achieve higher LDS scores. The Datamodels framework was optimized for this objective and trained as a supervised learning task, using tens of thousands of models. Hence, it achieves a better correlation with LDS. 

It is important to highlight that while Datamodels and TRAK outperform our baseline in terms of LDS with extensive model ensembles, this metric provides limited insights into understanding machine learning models. Our baseline approach excels in data brittleness metrics, offering a faithful representation of which training samples provide the most positive influence for a test sample.

\end{document}